\documentclass[10pt,twocolumn,letterpaper]{article}

\usepackage[pagenumbers]{wacv} 
\usepackage[ruled,vlined]{algorithm2e}
\usepackage[table]{xcolor}
\definecolor{deepgreen}{RGB}{0,100,0}
\usepackage{makecell}
\usepackage{graphicx}
\usepackage[percent]{overpic}
\usepackage{tabularx}
\usepackage{float}

\definecolor{wacvblue}{rgb}{0.21,0.49,0.74}
\usepackage[pagebackref,breaklinks,colorlinks,allcolors=wacvblue]{hyperref}

\def\confName{WACV}
\def\confYear{2026}

\title{HIVTP: A Training-Free Method to Improve VLMs Efficiency via Hierarchical Visual Token Pruning Using Middle-Layer-Based Importance Score}

\author{
Jingqi Xu\thanks{Equal contribution.} \quad
Jingxi Lu\footnotemark[1] \quad
Chenghao Li \quad
Sreetama Sarkar \quad
Peter A. Beerel \\
University of Southern California, Los Angeles, USA \\
{\tt\small \{jingqixu, jingxil, cli78217, sreetama, pabeerel\}@usc.edu}
}

\begin{document}
\maketitle
\begin{abstract}
Vision-Language Models (VLMs) have shown strong capabilities on diverse multimodal tasks. However, the large number of visual tokens output by the vision encoder severely hinders inference efficiency, and prior studies have shown that many of these tokens are not important and can therefore be safely pruned. 
In this work, we propose (\textbf{HIVTP}), a training-free method to improve VLMs efficiency via hierarchical visual token pruning using a novel middle-layer-based importance score. Specifically, we utilize attention maps extracted from the middle layers of the vision encoder, which better reflect fine-grained and object-level attention, to 
estimate
visual token importance.
Based on this,
we propose a hierarchical visual token pruning method to retain both globally and locally important visual tokens. 
Specifically, we reshape the 1-D visual token sequence output by the vision encoder into a 2-D spatial layout. In the global retaining stage, we divide the image into regions and retain tokens with higher importance scores in each region; in the local retaining stage, we then divide the image into small windows and retain the most important token in each local window.
Experimental results show that our proposed method, \textbf{HIVTP}, can reduce the time-to-first-token (TTFT) of LLaVA-v1.5-7B and LLaVA-Next-7B by up to 50.0\% and 55.1\%, respectively, and improve the token generation throughput by up to 60.9\% and 47.3\%, without sacrificing accuracy, and even achieving improvements on certain benchmarks. Compared with prior works, \textbf{HIVTP} achieves better accuracy while offering higher inference efficiency. \textbf{Code}: \url{https://github.com/Blacktower27/HIVTP}.

\end{abstract}
    
\section{Introduction}
\label{sec:intro}

Vision-Language Models (VLMs) such as Gemini Pro~\cite{comanici2025gemini}, GPT-5~\cite{openai2025gpt5}, and LLaVA~\cite{liu2024improvedbaselinesvisualinstruction, liu2024llavanext} have demonstrated strong capabilities across a wide range of vision-language tasks. 
They leverage a vision encoder to convert the input image into a sequence of visual tokens, which are then projected into the language feature space. Specifically, these visual tokens are concatenated with textual tokens and jointly processed by a Large Language Model (LLM)~\cite{vicuna2023} to generate responses. Typically, the number of visual tokens is significantly larger than that of textual tokens~\cite{zhang2025adaptivevisualtokenpruning, li2025inferenceoptimalvlmsneed, li2025windowtokenconcatenationefficient}. For instance, LLaVA-v1.5~\cite{liu2024improvedbaselinesvisualinstruction} converts an input image into 576 visual tokens using CLIP ViT-L/336~\cite{radford2021learningtransferablevisualmodels}, which is usually more than ten times the number of textual tokens~\cite{li2025windowtokenconcatenationefficient}. This large number of visual tokens hinders the inference efficiency of the LLM~\cite{zhang2025textvisualattentionexploitingvisual, zhang2025sparsevlmvisualtokensparsification, jiang2024foprufocalpruningefficient}, and prior studies~\cite{jiang2024foprufocalpruningefficient, li2025todrevisualtokenpruning, yang2024visionziplongerbetternecessary} have shown that many of these tokens are 
unimportant for the task at hand and can be removed while maintaining the performance of the VLMs.

Researchers have developed a variety of token compression methods~\cite{li2023blip, bai2023qwenvlversatilevisionlanguagemodel, li2025tokenpacker, cha2024honeybee} 
that leverage the attention mechanism~\cite{vaswani2023attentionneed} or convolution operation~\cite{oshea2015introductionconvolutionalneuralnetworks} to aggregate the visual information from a large number of visual tokens into a few learnable queries. Although these methods reduce the number of visual tokens while preserving the main visual information, they generally require training additional attention layers or convolutional layers, which demands significant additional computational resources.

Several training-free methods have also been proposed~\cite{yang2024visionziplongerbetternecessary, jiang2024foprufocalpruningefficient, shang2024llavaprumergeadaptivetokenreduction, arif2024hiredattentionguidedtokendropping}. 
Generally, these methods leverage the attention maps from the later layers of the vision encoder to estimate the importance of visual tokens, retain the more important ones, and either discard the less important ones or compress them into a few visual representations.
However, a recent study~\cite{10.1145/3664647.3681712} has shown that in the later layers of the vision encoder, the distribution of attention scores among visual tokens is highly imbalanced: only those encoding abstract high-level semantics receive high attention scores. As a result, existing methods that rely on attention maps from later layers to retain important visual tokens may lose fine-grained visual information, thereby degrading the performance of VLMs on tasks that require detailed visual understanding.

To address the aforementioned limitations, we propose \textbf{HIVTP}, a training-free method to improve VLMs efficiency by hierarchically pruning visual tokens using middle-layer-based importance scores. Specifically, 
we leverage the attention maps from the middle layers of the vision encoder, which have been shown in a recent study~\cite{10.1145/3664647.3681712} to better reflect object-level, fine-grained importance, to compute the importance scores of visual tokens. Then, we propose a hierarchical visual token pruning strategy to retain both globally and locally important visual tokens. 
We first reshape the 1-D visual token sequence output by the vision encoder into a 2-D spatial layout corresponding to the original image shape.
In the global retaining stage, motivated by the insight from~\cite{ye2024atpllavaadaptivetokenpruning, wen2025tokenpruningmultimodallarge} that uneven spatial distributions of retained visual tokens after pruning can degrade the performance of VLMs, we evenly divide the image into  regions and retain the tokens with higher importance scores from each region. In the local retaining stage, we divide the image into smaller windows and select the token with the highest importance score from the remaining tokens in each window.
To evaluate the effectiveness of our method, we apply it to two pioneering VLMs, LLaVA-v1.5-7B~\cite{liu2024improvedbaselinesvisualinstruction} and LLaVA-Next-7B~\cite{liu2024llavanext}, and conduct experiments on widely used benchmarks~\cite{fu2024mmecomprehensiveevaluationbenchmark, li2023evaluatingobjecthallucinationlarge, singh2019vqamodelsread, liu2024mmbenchmultimodalmodelallaround, lu2022learnexplainmultimodalreasoning, Liu_2024, ainslie2023gqatraininggeneralizedmultiquery, kembhavi2016diagramworthdozenimages}. Our experimental results show that \textbf{HIVTP} significantly improves the inference efficiency of both models without sacrificing much accuracy, and even achieves accuracy gains on certain benchmarks. Furthermore, compared with prior methods, \textbf{HIVTP} achieves higher accuracy while offering better inference efficiency overall.

Our main contributions are summarized as follows: 1) We propose \textbf{HIVTP}, a training-free method that improves the efficiency of VLMs with minimal accuracy loss. 
2) We introduce a novel middle-layer-based importance scoring strategy to measure the importance of visual tokens.
3) 
We propose a hierarchical visual token pruning strategy that retains both globally and locally important visual tokens.
4) 
Extensive experiments on widely used benchmarks and various VLMs show that \textbf{HIVTP} reduces the time-to-first-token (TTFT) by up to 26.5\% and increases the token generation throughput by up to 11.1\% compared with prior methods, demonstrating the superiority of \textbf{HIVTP} over prior methods in improving inference efficiency while preserving accuracy.

\section{Related Work}



\subsection{Vision-Language Models (VLMs)}

Closed-source Vision-Language Models (VLMs), such as Gemini Pro~\cite{comanici2025gemini, team2024gemini} and GPT-5~\cite{openai2025gpt5}, have demonstrated remarkable progress in multimodal understanding, achieving strong performance across a wide range of vision-language tasks. In parallel, the open-source community has also advanced rapidly. Representative models, including the BLIP2~\cite{li2023blip}, MiniGPT-4~\cite{zhu2023minigpt}, and QwenVL~\cite{bai2023qwenvlversatilevisionlanguagemodel}, leverage the Q-Former~\cite{li2023blip} as a projector to compress visual tokens and align vision-text features. In contrast, the LLaVA series~\cite{liu2024improvedbaselinesvisualinstruction, liu2024llavanext} and more recent frameworks, such as VILA~\cite{lin2024vilapretrainingvisuallanguage}, ShareGPT4V~\cite{chen2024sharegpt4v}, CogVLM~\cite{wang2024cogvlm}, DeepSeekVL~\cite{lu2024deepseek}, and InternVL~\cite{chen2024internvl}, adopt lightweight MLP-based projectors to connect vision and language models. Collectively, these developments highlight the rapid evolution of VLMs and their increasing capacity to tackle complex multimodal 
tasks.


\subsection{Visual Token Pruning}
Despite the strong performance of current VLMs across various multimodal tasks, their inference efficiency is hindered by the large number of visual tokens. To address this problem, several visual token pruning methods have been proposed, which can be broadly categorized into two classes: training-based methods and training-free methods. Within the category of training-based methods, Q-Former~\cite{li2023blip} and Resampler~\cite{bai2023qwenvlversatilevisionlanguagemodel} train additional cross-attention layers to aggregate the information from a large number of visual tokens into a few learnable queries. TokenPacker~\cite{li2025tokenpacker} first downsamples visual tokens into a few queries and then integrates the information from the original visual tokens into these queries through an attention mechanism. Honeybee~\cite{cha2024honeybee} aggregates local visual tokens into a single visual token using either convolution or attention mechanisms.
However, these training-based methods require training additional components, thus leading to substantial computational resource consumption. FoPru~\cite{jiang2024foprufocalpruningefficient} and HiRED~\cite{arif2024hiredattentionguidedtokendropping} are training-free methods that utilize the multi-head attention maps from the penultimate layer of the vision encoder to compute the average attention score for each visual token as its importance score, and prune the less important ones.
Building upon similar importance score computing strategies, VisionZip~\cite{yang2024visionziplongerbetternecessary} and PruMerge~\cite{shang2024llavaprumergeadaptivetokenreduction} perform average merging among less important visual tokens and preserve the resulting merged representations. However, these existing training-free methods that rely on attention maps from the later layers of the vision encoder to select important visual tokens may lose fine-grained visual information, because the recent study~\cite{10.1145/3664647.3681712} has shown that the distribution of attention scores among visual tokens in the later layers is highly imbalanced, with only those encoding abstract high-level semantics receive high attention scores.

\begin{figure*}[t]
    \centering
    \includegraphics[width=1\textwidth]{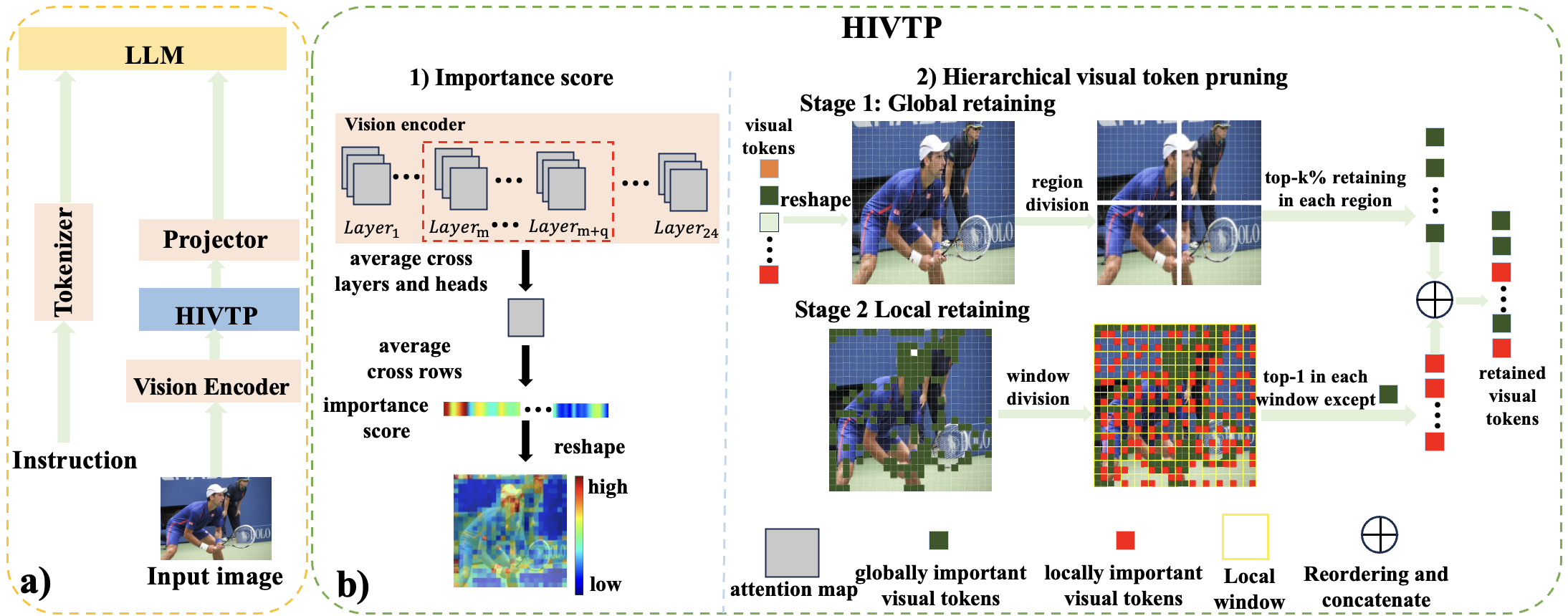}
    \caption{Framework of HIVTP. a) The overall workflow of HIVTP for VLMs. b) First, we leverage the attention maps from the middle layers of the vision encoder to compute the importance scores of visual tokens. Then, we adopt a hierarchical visual token pruning strategy to retain both globally important and locally important visual tokens.}
    \label{fig:workflow}
\end{figure*}

\section{Preliminaries}
\label{sec:VLMs architecture}
VLMs generate textual responses 
based on an image–instruction pair~\cite{jiang2024foprufocalpruningefficient, liu2023visualinstructiontuning}. 
They typically comprise three components: a vision encoder, a projector, and a LLM~\cite{vicuna2023}.
The vision encoder first partitions the input image into $N = n \times n$ non-overlapping patches.
Each patch is linearly mapped into a $d_v$-dimensional embedding. 
These patch embeddings are arranged into a sequence. To this sequence, a learnable class token $\mathrm{CLS} \in \mathbb{R}^{1 \times d_v}$ is prepended to facilitate global visual information aggregation, resulting in the visual token matrix $E_v \in \mathbb{R}^{(N+1) \times d_v}$.
These $N+1$ visual tokens are subsequently processed by 
a set of transformer layers~\cite{vaswani2023attentionneed} within the vision encoder to facilitate visual information interaction. We only pass the $N$ output tokens related to the original image from the penultimate layer of the vision encoder\footnote{The final layer of the vision encoder, in fact, is not used and can be omitted from the computation~\cite{shang2024llavaprumergeadaptivetokenreduction}.}, referred to as $Z_v \in \mathbb{R}^{N \times d_v}$, to the subsequent projector.
The projector aligns the visual and textual modalities by mapping $Z_v$ into the language feature space, yielding $V \in \mathbb{R}^{N \times d}$, where $d$ denotes the shared embedding dimensionality between modalities.
Meanwhile, the instruction is tokenized and embedded into textual embeddings
$T \in \mathbb{R}^{M \times d}$, 
where $M$ denotes the number of textual tokens.
Finally, the visual tokens $V$ and the textual tokens $T$ are concatenated and fed into the LLM, which performs cross-modal interactions through a set of transformer layers and autoregressively generates a textual response $R$ of length $L_R$,
\begin{equation}
P(R \mid V, T) \;=\; \prod_{i=1}^{L_R} P\!\left(R_i \mid V, T, R_{<i}\right).
\end{equation}
\noindent
The prediction of each response token is conditioned on the visual tokens $V$, the textual tokens $T$, and previously generated tokens in the response. Consequently, the inference efficiency is heavily affected by the total sequence length, with the number of visual tokens $N$ often far exceeding the number of textual tokens $M$~\cite{zhang2025adaptivevisualtokenpruning, li2025inferenceoptimalvlmsneed, li2025windowtokenconcatenationefficient, wen2024efficientvisionlanguagemodelssummarizing, li2025todrevisualtokenpruning}. 
Therefore, pruning less informative visual tokens is an effective strategy for improving inference efficiency.
\section{HIVTP}
\label{sec:Methodology}

\subsection{Overview}
The framework of HIVTP is illustrated in Figure~\ref{fig:workflow}. HIVTP identifies and retains important visual tokens from the output of the penultimate layer of the vision encoder. 
We first estimate the importance of visual tokens based on the attention maps from the middle layers of the vision encoder. Then, we adopt a hierarchical visual token pruning strategy to retain both globally and locally important visual tokens. These retained tokens are then projected into the language feature space.
\subsection{Visual Token Importance Score}
\label{sec:importance score}
To prune 
unimportant visual tokens, we first estimate the importance of each token. 
A common strategy for estimating visual token importance is to measure their average attention scores using multi-head attention maps extracted from selected layers of the vision encoder~\cite{jiang2024foprufocalpruningefficient}.

While previous methods~\cite{yang2024visionziplongerbetternecessary, jiang2024foprufocalpruningefficient, shang2024llavaprumergeadaptivetokenreduction, arif2024hiredattentionguidedtokendropping}  typically select later layers of the vision encoder for this purpose, a recent study~\cite{10.1145/3664647.3681712} has shown that different vison encoder layers focus on different levels of visual semantics. Specifically, in 
earlier layers, each visual token tends to receive a similar amount of attention from other visual tokens. In middle layers, visual tokens associated with object-level semantics tend to receive higher attention from other visual tokens, whereas in later layers, tokens encoding abstract high-level visual semantics are more strongly attended. As a result, using multi-head attention maps from later layers may overlook fine-grained, object-level visual information, which can degrade performance on tasks that require detailed visual understanding. 
In addition, we visualize several examples from the MME benchmark~\cite{fu2024mmecomprehensiveevaluationbenchmark}, showing the attention heatmaps at different layers of the vision encoder in LLaVA-v1.5-7B~\cite{liu2024improvedbaselinesvisualinstruction}, with brighter tokens indicating higher attention scores. We show one example in Figure~\ref{fig:attention_heatmaps}, and additional examples are provided in Appendix~\ref{subsec:heatmap}. 
This further verifies the behavioral characteristics of the vision encoder across different layer depths. 


To avoid performance degradation on fine-grained visual tasks, we capture object-level visual information by extracting multi-head attention maps from a set of middle layers in the vision encoder.
Specifically, the vision encoder comprises $L$ layers, indexed by $\mathcal{I}(L) = \{1, \dots, L\}$ , with each layer contain $H$ attention heads. We denote the complete collection of multi-head attention maps from all vision encoder layers as,
\[
\mathbf{A} = \left\{ \mathbf{A}^{(l,H)} \in \mathbb{R}^{H\times(N+1) \times (N+1)} \,\middle|\, l \in \mathcal{I}(L)\right\},
\]
where $\mathbf{A}^{(l,H)}$ represents $H$ attention maps produced in layer $l$. We define the index set of selected middle layers as $L_s = \{l_m, \cdots, l_{m+q}\} \subseteq \mathcal{I}(L)$, which is 
identified in Section~\ref{sec: setup}. 
We extract attention maps from each selected layer $l \in L_s$, and define the importance score of each visual token as the average attention it receives from other visual tokens,



\begin{equation}
\bar{\mathbf{A}}^{(H)} = \frac{1}{|L_s|}\sum_{l \in L_s} \mathbf{A}^{(l,H)}, 
\end{equation}

\begin{equation}
\bar{\mathbf{A}} = \frac{1}{H}\sum_{h=1}^{H} \bar{\mathbf{A}}^{(H)}, 
\end{equation}

\begin{equation}
\mathbf{S} = \frac{1}{N+1}\sum_{j=0}^{N} \bar{\mathbf{A}}_{j,1:} 
\end{equation}
where $\bar{\mathbf{A}} \in \mathbb{R}^{(N+1) \times (N+1)}$ and
$\bar{\mathbf{A}}_{j,1:}$ is the $j^{th}$ row of $\bar{\mathbf{A}}$ excluding the element corresponding to the CLS token. 
As a result, $\mathbf{S} \in \mathbb{R}^{N \times 1}$ represents the averaged importance scores of the $N$ visual tokens. 
Based on $\mathbf{S}$, we apply a hierarchical visual token pruning strategy to select both globally and locally important visual tokens, as described below.

\subsection{Hierarchical Visual Token Pruning}
\label{sec:Hierarchical}

Our hierarchical visual token pruning strategy aims to retain important object-level visual tokens in a global-to-local manner through two sequential stages. We first extract the output visual tokens $\mathbf{Z}_v \in \mathbb{R}^{N \times d_v}$ from the penultimate layer of the vision encoder and compute their corresponding importance scores $\mathbf{S} \in \mathbb{R}^{N \times 1}$ based on the method described in Section~\ref{sec:importance score}. Subsequently, we perform a global retaining stage and a local retaining stage to retain visual tokens that are globally and locally important, respectively.

\noindent\textbf{Global Retaining Stage.}
We aim to select the top-$k\%$ most important globally informative visual tokens from the $N$ visual tokens $\mathbf{Z}_v \in \mathbb{R}^{N \times d_v}$. Previous studies~\cite{ye2024atpllavaadaptivetokenpruning, wen2025tokenpruningmultimodallarge} have validated that uneven spatial distributions of preserved visual tokens after pruning can degrade the performance of VLMs. 
Therefore, it is necessary to keep the retained visual tokens spatially uniform to maintain the model's accuracy. Based on this insight, we uniformly partition the visual tokens into multiple large regions, and perform regional top-$k\%$ selection to enhance the spatial uniformity of the retained tokens.

Specifically, we reshape the visual tokens $\mathbf{Z}_v \in \mathbb{R}^{N \times d_v}$ into a 2-D structure $\mathbf{Z}'_v \in \mathbb{R}^{n \times n \times d_v}$ that reflects the relative spatial positions within the original image. 
Then, we uniformly divide 
$\mathbf{Z}'_v$ into $r^2$ non-overlapping regions and denote the set of regions as $\{\mathcal{R}_i\}_{i=1}^{r^2}$.  
Each region $\mathcal{R}_i$ contains $N_r = \left( \frac{n}{r} \right)^2$ visual tokens, and we let 
$\mathcal{I}(\mathcal{R}_i) \subseteq \{1, \dots, N\}$ denote their positional indices with respect to the original $\mathbf{Z}_v$.
For each region $\mathcal{R}_i$, we extract the corresponding importance scores as $\mathbf{S}_i = \mathbf{S}[\mathcal{I}(\mathcal{R}_i)]$.
We then select the indices of the top-$k\%$ visual tokens with the highest importance scores in this region based on $\mathbf{S}_i$,
\begin{equation}
\mathcal{I}^{\text{global}}_i = \texttt{TopK}(\mathcal{I}(\mathcal{R}_i), \mathbf{S}_i, k\%).
\end{equation}
where $\texttt{TopK}(\cdot)$ returns the subset of $\mathcal{I}(\mathcal{R}_i)$ corresponding to the top-$k\%$ indices with the highest importance scores given $\mathbf{S}_i$, and $I^{\text{global}}_i$ denotes the index set of globally important visual tokens retained from region $\mathcal{R}_i$.
The union of the index sets of globally important visual tokens retained from all regions forms the global visual token set $\mathcal{I}^{\text{global}}$,
\begin{equation}
\mathcal{I}^{\text{global}} = \bigcup_{i=1}^{r^2} \mathcal{I}^{\text{global}}_i.
\label{eq:global_token_union}
\end{equation}
We denote the total number of retained globally important visual tokens as $p_g = |\mathcal{I}^{\text{global}}|$, where $|\mathcal{I}^{\text{global}}| = r^2 \cdot N_r \cdot k\%$. Through the global retaining stage, we thus retain globally important visual tokens while ensuring a form of global spatial uniformity. 

\begin{figure}[t]
    \centering
    \includegraphics[width=0.85\linewidth]{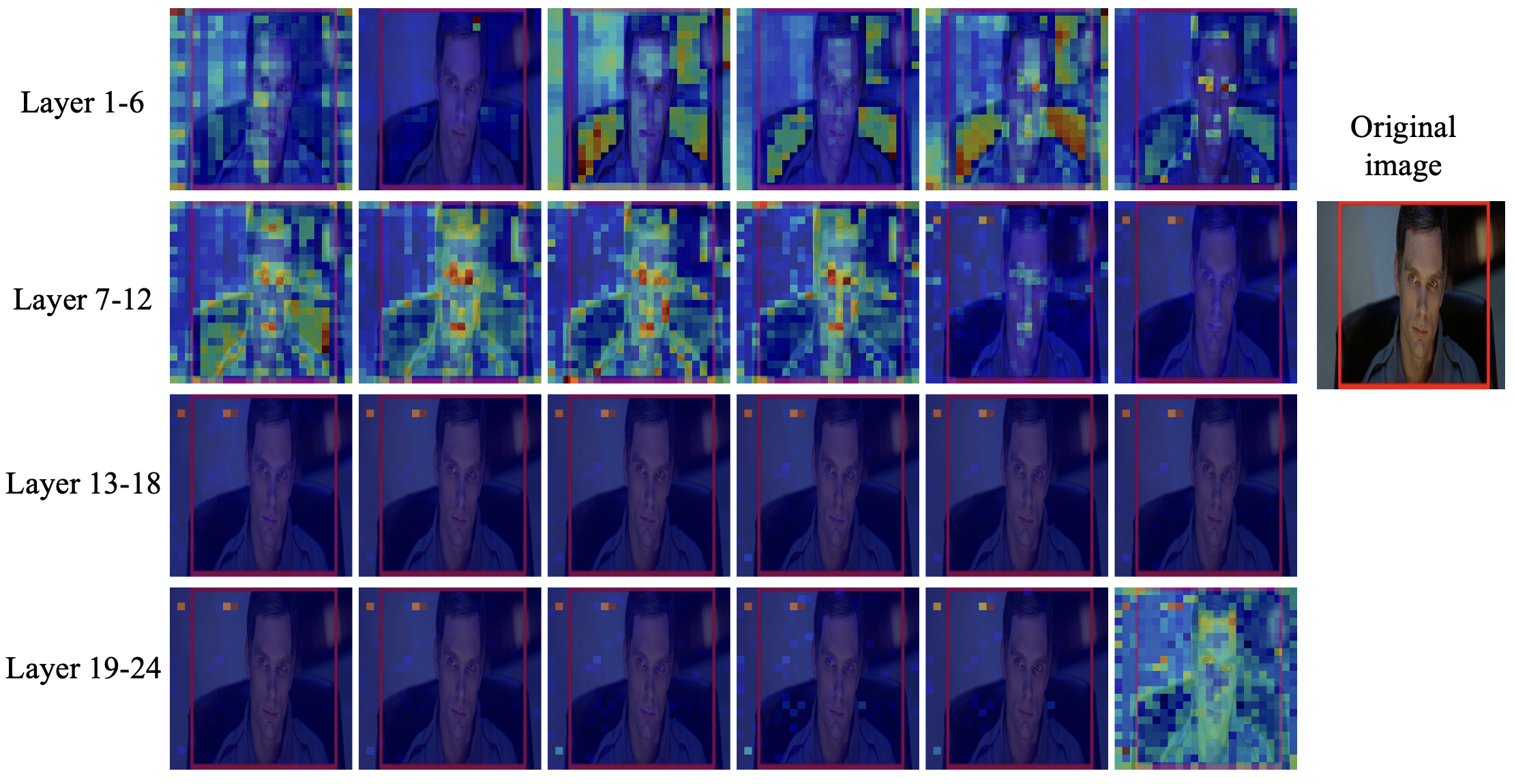}
    \caption{Attention heatmaps at different layers of the vision encoder in LLaVA-v1.5-7B for one example from the MME benchmark. From layer 7 to layer 10, the visual tokens corresponding to the main objects in the image exhibit higher brightness, indicating higher attention weights.}
    \label{fig:attention_heatmaps}
\end{figure}

\noindent\textbf{Local Retaining Stage.}
Although the global retaining stage identifies globally important visual tokens via region division followed by top-$k\%$ retaining, locally important fine-grained visual tokens may be overlooked. 
To address this, in the local retaining stage, we explicitly perform local spatial modeling to identify locally important visual tokens. 

Specifically, we partition $\mathbf{Z}'_v \in \mathbb{R}^{n \times n \times d_v}$ into multiple non-overlapping local windows of size $c \times c$, where $c$ denotes the number of visual tokens along both the height and width within each window, and $n$ is assumed to be divisible by $c$. As a result, the total number of such windows is denoted by $N_{\mathrm{w}} = \left(\frac{n}{c}\right)^2$, and we denote the set of all windows as $\{\mathcal{W}_j\}_{j=1}^{N_{\mathrm{w}}}$.
To identify locally important visual tokens, we denote the set of positional indices of visual tokens in window $\mathcal{W}_j$ with respect to $\mathbf{Z}_v$ as $\mathcal{I}(\mathcal{W}_j) \subseteq \{1, \dots, N\}$.  
To avoid repeated selections, we exclude tokens that have already been retained in the global stage and define the remaining candidates as $\mathcal{U}_j = \mathcal{I}(\mathcal{W}_j) \setminus \mathcal{I}^{\text{global}}$.
If $\mathcal{U}_j$ is non-empty, we collect the index of the visual token with the highest importance score as the locally important visual token in window $\mathcal{W}_j$. Otherwise, no locally important visual token is retained from $\mathcal{W}_j$,
\begin{equation}
\mathcal{I}^{\text{local}}_j =
\begin{cases}
\arg\max\limits_{t \in \mathcal{U}_j} \mathbf{S}[t], & \text{if } \mathcal{U}_j \neq \emptyset, \\
\emptyset, & \text{otherwise}
\end{cases}
\quad \forall j \in \{1, \dots, N_{\mathrm{w}}\}.
\label{eq:local_selection}
\end{equation}
where $I^{\text{local}}_j$ denotes the index set of the locally important visual token retained from the window $\mathcal{W}_j$.


The union of the index sets of locally important visual tokens retained from all windows forms the locally visual token set $\mathcal{I}^{\text{local}}$,
\begin{equation}
\mathcal{I}^{\text{local}} = \bigcup_{j=1}^{N_{\mathrm{w}}} \mathcal{I}^{\text{local}}_j.
\label{eq:local_set}
\end{equation}
We denote the length of $\mathcal{I}^{\text{local}}$ as $p_l$.
Since some windows may contribute no tokens (i.e., $\mathcal{U}_j = \emptyset$), it follows that $p_l \leq N_{\mathrm{w}}$.

The index sets of retained globally and locally important visual tokens, $\mathcal{I}^{\text{global}}$ and $\mathcal{I}^{\text{local}}$, are combined to form the final index set $\mathcal{I}^{\text{final}}$ of retained visual tokens,
\begin{equation}
\mathcal{I}^{\text{final}} = \mathcal{I}^{\text{global}} \cup \mathcal{I}^{\text{local}}.
\label{eq:final_index_union}
\end{equation}

To preserve spatial consistency and maintain the semantic coherence of retained visual tokens, we sort $\mathcal{I}^{\text{final}}$ in ascending order to obtain the sorted index list $\mathcal{I}^{\text{final}}_{\text{sorted}}$, and extract the corresponding visual tokens from $\mathbf{Z}_v \in \mathbb{R}^{N \times d_v}$ to form the retained 
list $\mathbf{Z}_v^{\text{retained}} \in \mathbb{R}^{p \times d_v}$, where $p = p_g + p_l$ is the total number of retained tokens.
We define the visual token retaining ratio $r_{\text{retain}}$ as,
\begin{equation}
r_{\text{retain}} = \frac{p}{N}.
\label{eq:retention_ratio}
\end{equation}
Since the number of retained locally important visual tokens satisfies $p_l \leq N_{\mathrm{w}}$, the total number of retained tokens is upper-bounded by $p_{\text{max}} = p_g + N_{\mathrm{w}},\label{eq:p_max}$ leading to $r_{\text{max}}$ as the upper bound of $r_{\text{retain}}$,
\begin{equation}
r_{\text{retain}} \leq r_{\text{max}} = \frac{p_{\text{max}}}{N}.
\label{eq:r_max}
\end{equation}
In practice, we treat $r_{\text{max}}$ as a hyperparameter to control the inference efficiency.


After hierarchical visual token pruning, the retained visual tokens $Z_v^{\text{retained}} \in \mathbb{R}^{p \times d_v}$ are projected to align with the textual modality and subsequently concatenated with the textual tokens. The resulting multimodal representation is then jointly fed into the LLM. 
The algorithmic details are presented in Appendix~\ref{subsec:alg}.

\begin{table*}[ht]
\centering
\begingroup
\scriptsize
\setlength{\tabcolsep}{2pt} 
\renewcommand{\arraystretch}{0.94} 
\setlength{\aboverulesep}{0.5pt}
\setlength{\belowrulesep}{0.5pt}
\renewcommand\cellgape{\relax}
\resizebox{0.98\textwidth}{!}{
\begin{tabular}{c|c|c|cccccccc|cc}
\toprule
& & & \multicolumn{8}{c|}{\textbf{Accuracy Performance}} & \multicolumn{2}{c}{\textbf{Efficiency}} \\
\textbf{Model \& Method} & \textbf{\large  $r_{retain}$} & \makecell{\# Visual \\ tokens} 
& MME~\cite{fu2024mmecomprehensiveevaluationbenchmark}~$\uparrow$ & POPE~\cite{li2023evaluatingobjecthallucinationlarge}~$\uparrow$ & $\mathrm{VQA}^{\text{T}}$~\cite{singh2019vqamodelsread}~$\uparrow$ 
& MMB~\cite{liu2024mmbenchmultimodalmodelallaround}~$\uparrow$ & SQA~\cite{lu2022learnexplainmultimodalreasoning}~$\uparrow$ & OCR~\cite{Liu_2024}~$\uparrow$ 
& GQA~\cite{ainslie2023gqatraininggeneralizedmultiquery}~$\uparrow$ & Ai2D~\cite{kembhavi2016diagramworthdozenimages}~$\uparrow$ 
& \makecell{TTFT\\(ms)~$\downarrow$} & \makecell{Throughput\\(token/second)~$\uparrow$} \\

\midrule
\rowcolor{gray!15}
\multicolumn{13}{c}{\textit{$r_{max}$ = 100\% (576 Visual Tokens)}} \\
LLaVA-v1.5-7B ~\cite{liu2024improvedbaselinesvisualinstruction} & 100.0\% & 576 & 1513.51 & 86.98 & 46 & 64.7 & 69.51 & 31.2 & 61.97 & 55.5 & 140 & 18.66 \\
\midrule
\rowcolor{gray!15}
\multicolumn{13}{c}{\textit{$r_{max}$ = 75\% (432 Visual Tokens)}} \\
FoPru~\cite{jiang2024foprufocalpruningefficient} & 75.0\% & 432 & 1508.98 & 86.51 & 45.40 & 64.18 & 70.02 & 31.00 & 53.41 & 54.07 & 121 & 21.95 \\
VisionZip~\cite{yang2024visionziplongerbetternecessary} & 75.0\% & 432 & 1497.65 & 86.82 & 41.08 & 64.00 & 70.10 & 29.80 & 59.48 & 54.92 & 119 & 22.49 \\
PruMerge~\cite{shang2024llavaprumergeadaptivetokenreduction} & 75.0\% & 432 & 1462.31 & 84.78 & 39.43 & 61.86 & \textbf{70.12} & 28.70 & 60.57 & 53.92 & 124 & 21.8 \\
HiRED~\cite{arif2024hiredattentionguidedtokendropping} & 75.0\% & 432 & 1502.73 & 86.93 & 45.27 & 63.40 & 69.58 & 30.90 & 60.60 & 54.40 & 117 & 22.75 \\
\makecell[c]{HIVTP (Ours) \\ $(r{=}2, k{=}50, c{=}2)$} & 70.1\% & 404  & \textbf{1512.41} & \textbf{87.20} & \textbf{45.44} & \textbf{65.03} & 69.89 & \textbf{31.40} & \textbf{61.42} & \textbf{54.98} & \textbf{114} & \textbf{22.95} \\
\midrule
\rowcolor{gray!15}
\multicolumn{13}{c}{\textit{$r_{max}$ = 50\% (288 Visual Tokens)}} \\
FoPru~\cite{jiang2024foprufocalpruningefficient} & 50.0\% & 288 & 1445.73 & 84.89 & 45.55 & 62.80 & 69.84 & 30.00 & 59.80 & 54.07 & 100 & 22.59 \\
VisionZip~\cite{yang2024visionziplongerbetternecessary} & 50.0\% & 288 & 1464.15 & 86.51 & 41.07 & \textbf{64.09} & 69.89 & 29.50 & 59.18 & 54.92 & 97 & 23.14 \\
PruMerge~\cite{shang2024llavaprumergeadaptivetokenreduction} & 50.0\% & 288 & 1432.62 & 85.13 & 39.53 & 61.77 & 69.89 & 29.40 & 60.49 & 53.72 & 101 & 22.39 \\
HiRED~\cite{arif2024hiredattentionguidedtokendropping} & 50.0\% & 288 & 1474.46 & 87.04 & 45.46 & 62.97 & 69.70 & 30.00 & 60.50 & 54.88 & 100 & 22.04 \\
\makecell[c]{HIVTP (Ours) \\ $(r{=}2, k{=}25, c{=}2)$} & 48.9\% & 282  & \textbf{1495.37} & \textbf{87.12} & \textbf{45.84} & 64.00 & \textbf{70.12} & \textbf{30.50} & \textbf{60.66} & \textbf{54.99} & \textbf{92} & \textbf{23.7} \\
\midrule
\rowcolor{gray!15}
\multicolumn{13}{c}{\textit{$r_{max}$ = 40\% (230 Visual Tokens)}} \\
FoPru~\cite{jiang2024foprufocalpruningefficient} & 40.0\% & 230 & 1445.51 & 87.01 & 44.91 & 61.32 & 69.37 & 29.40 & 59.87 & 54.47 & 75 & 26.71 \\
VisionZip~\cite{yang2024visionziplongerbetternecessary} & 40.0\% & 230 & 1451.11 & 86.79 & 41.43 & 62.12 & 69.72 & 29.60 & 58.94 & 54.76 & 102 & 25.20 \\
PruMerge~\cite{shang2024llavaprumergeadaptivetokenreduction} & 40.0\% & 230 & 1444.13 & 85.23 & 39.64 & 61.77 & 69.63 & 28.40 & 59.32 & 54.92 & 90 & 26.80 \\
HiRED~\cite{arif2024hiredattentionguidedtokendropping} & 40.0\% & 230 & 1442.41 & 86.83 & \textbf{45.22} & 61.49 & 69.75 & 30.10 & 59.94 & 54.92 & 80 & 27.94 \\
\makecell[c]{HIVTP (Ours) \\ $(r{=}2, k{=}15, c{=}2)$} & 39.2\% & 226  & \textbf{1454.26} & \textbf{87.07} & 45.04 & \textbf{62.20} & \textbf{70.01} & \textbf{30.20} & \textbf{60.20} & \textbf{55.02} & \textbf{75} & \textbf{27.99} \\
\midrule
\rowcolor{gray!15}
\multicolumn{13}{c}{\textit{$r_{max}$ = 25\% (144 Visual Tokens)}} \\
FoPru~\cite{jiang2024foprufocalpruningefficient} & 25.0\% & 144 & 1400.14 & 85.01 & 44.25 & 61.86 & 69.72 & \textbf{30.50} & 57.44 & 54.73 & 71 & 29.91 \\
VisionZip~\cite{yang2024visionziplongerbetternecessary} & 25.0\% & 144 & 1417.35 & 85.04 & 41.09 & 60.74 & 69.70 & 28.00 & 57.05 & 54.16 & 72 & 29.71 \\
PruMerge~\cite{shang2024llavaprumergeadaptivetokenreduction} & 25.0\% & 144 & 1400.91 & 84.37 & 38.80 & 60.22 & 69.77 & 28.00 & 58.06 & 54.37 & 74 & 29.46 \\
HiRED~\cite{arif2024hiredattentionguidedtokendropping} & 25.0\% & 144 & \textbf{1436.28} & 85.13 & 44.24 & 61.43 & 69.65 & 28.40 & 58.31 & 54.89 & 77 & 28.89 \\
\makecell[c]{HIVTP (Ours) \\ $(r{=}2, k{=}14, c{=}3)$} & 24.6\% & 142 & 1403.48 & \textbf{86.80} & \textbf{44.40} & \textbf{61.99} & \textbf{69.94} & 28.80 & \textbf{58.70} & \textbf{54.92} & \textbf{70} & \textbf{30.02} \\
\bottomrule
\end{tabular}}
\endgroup
\caption{Accuracy and efficiency comparison of HIVTP with prior methods on LLaVA-v1.5-7B under different upper bounds of the visual token retaining ratio, $r_{\text{max}} \in \{75\%, 50\%, 40\%,25\%\}$. The actual retaining ratio is denoted by $r_{\text{retain}}$. TTFT and Throughput are computed as averages across all benchmarks. The best results among all methods for each $r_{\text{max}}$ are highlighted in \textbf{bold}.}
\label{tab:main_llava15}
\end{table*}

\section{Experimental Results}
\label{sec:Experiment}
In this section, we systematically evaluate our proposed HIVTP on eight different benchmarks when applied to LLaVA-v1.5-7B~\cite{liu2024improvedbaselinesvisualinstruction} and LLaVA-Next-7B~\cite{liu2024llavanext}, and compare it with prior methods: FoPru~\cite{jiang2024foprufocalpruningefficient}, VisionZip~\cite{yang2024visionziplongerbetternecessary}, PruMerge~\cite{shang2024llavaprumergeadaptivetokenreduction}, and HiRED~\cite{arif2024hiredattentionguidedtokendropping}. 
\subsection{Experimental Setup}
\label{sec: setup}
We adopt eight widely used benchmarks, including: 1) MME~\cite{fu2024mmecomprehensiveevaluationbenchmark}, for evaluating visual perception and recognition abilities; 2) POPE~\cite{li2023evaluatingobjecthallucinationlarge}, for assessing the model's ability to avoid object hallucination; 3) TextVQA ($\mathrm{VQA}^{\text{T}}$) ~\cite{singh2019vqamodelsread}, for evaluating visual reasoning ability based on textual content in images; 4) MMBench (MMB)~\cite{liu2024mmbenchmultimodalmodelallaround}, for evaluating multimodal understanding and reasoning capabilities; 5) ScienceQA (SQA)~\cite{lu2022learnexplainmultimodalreasoning}, for evaluating the ability to answer scientific questions across multiple domains using image content; 6) OCRBench (OCR)~\cite{Liu_2024}, for recognizing and reasoning about text within images; 7) GQA~\cite{ainslie2023gqatraininggeneralizedmultiquery}, for evaluating the ability to answer questions about real-world scene graphs. and 8) Ai2D~\cite{kembhavi2016diagramworthdozenimages}, for evaluating the ability to understand and answer questions about scientific diagrams.
We conduct experiments under different upper bounds of the visual token keeping ratio, $r_{\text{max}} \in \{75\%, 50\%, 40\% ,25\%\}$. For LLaVA-Next, the original input image is divided into several parts as sub-images. Each sub-image, as well as the original image itself, is independently processed by the vision encoder to obtain its own visual tokens. We apply HIVTP to the visual tokens from both the original image and all sub-images. For prior methods, we set their actual visual token keeping ratio $r_{\text{keep}}$ to match $r_{\text{max}}$. All experiments are conducted using an NVIDIA A40 GPU.

\noindent\textbf{Middle Layers Setting.} 
To determine the index set of middle layers $L_s$ used for computing visual token importance scores in our method, we visualize the attention heatmaps at different layers of the vision encoder in LLaVA-v1.5-7B for several examples from the MME benchmark. One representative visualization is shown in Figure~\ref{fig:attention_heatmaps}, and additional examples are provided in Appendix~\ref{subsec:heatmap}.
As observed, from layer 7 to layer 10, the visual tokens corresponding to the main objects in the image exhibit higher brightness, indicating that they receive higher attention. Based on this observation, 
We select layers 7 to layer 10 as the middle layers.

\begin{table*}[ht]
\centering
\begingroup
\scriptsize
\setlength{\tabcolsep}{2pt} 
\renewcommand{\arraystretch}{0.94} 
\setlength{\aboverulesep}{0.5pt}
\setlength{\belowrulesep}{0.5pt}
\renewcommand\cellgape{\relax}
\resizebox{0.98\textwidth}{!}{
\begin{tabular}{c|c|c|cccccccc|cc}
\toprule
& & & \multicolumn{8}{c|}{\textbf{Accuracy Performance}} & \multicolumn{2}{c}{\textbf{Efficiency}} \\
\textbf{Model \& Method} & \textbf{\large  $r_{retain}$} & \makecell{\# Visual \\ tokens} 
& MME~\cite{fu2024mmecomprehensiveevaluationbenchmark}~$\uparrow$ & POPE~\cite{li2023evaluatingobjecthallucinationlarge}~$\uparrow$ & $\mathrm{VQA}^{\text{T}}$~\cite{singh2019vqamodelsread}~$\uparrow$ 
& MMB~\cite{liu2024mmbenchmultimodalmodelallaround}~$\uparrow$ & SQA~\cite{lu2022learnexplainmultimodalreasoning}~$\uparrow$ & OCR~\cite{Liu_2024}~$\uparrow$ 
& GQA~\cite{ainslie2023gqatraininggeneralizedmultiquery}~$\uparrow$ & Ai2D~\cite{kembhavi2016diagramworthdozenimages}~$\uparrow$ 
& \makecell{TTFT\\(ms)~$\downarrow$} & \makecell{Throughput\\(token/second)~$\uparrow$} \\

\midrule
\rowcolor{gray!15}
\multicolumn{13}{c}{\textit{$r_{max}$ = 100\% (2614 Visual Tokens)}} \\
LLaVA-Next-7B~\cite{liu2024llavanext} & 100.0\% & 2614 & 1484.20 & 87.74 & 64.78 & 67.87 & 78.19 & 50.2 & 64.35 & 66.32 & 423 & 13.28 \\
\midrule
\rowcolor{gray!15}
\multicolumn{13}{c}{\textit{$r_{max}$ = 75\% (1959 Visual Tokens)}} \\
FoPru~\cite{jiang2024foprufocalpruningefficient} & 75.0\% & 1959 & 1492.66 & 87.99 & 64.73 & 67.35 & 78.00 & 50.60 & 64.09 & 66.06 & 353 & 14.77 \\
VisionZip~\cite{yang2024visionziplongerbetternecessary} & 75.0\% & 1959 & 1479.59 & 87.99 & 64.45 & 67.53 & 78.00 & 50.00 & 64.10 & 65.93 & 343 & 14.88 \\
PruMerge~\cite{shang2024llavaprumergeadaptivetokenreduction} & 75.0\% & 1959 & 1270.49 & 79.53 & 50.79 & 62.80 & 78.68 & 39.20 & 61.72 & 63.41 & 344 & 15.09 \\
HiRED~\cite{arif2024hiredattentionguidedtokendropping} & 75.0\% & 1959 & 1478.38 & 88.01 & 64.50 & 67.53 & 78.09 & \textbf{50.80} & 64.27 & 66.13 & 345 & 14.96 \\
\makecell[c]{HIVTP (Ours) \\ $(r{=}2, k{=}50, c{=}2)$} & 71.7\% & 1857 & \textbf{1509.23} & \textbf{88.46} & \textbf{64.75} & \textbf{67.99} & \textbf{78.95} & 49.30 & \textbf{64.31} & \textbf{66.89} & \textbf{324} & \textbf{15.47} \\
\midrule
\rowcolor{gray!15}
\multicolumn{13}{c}{\textit{$r_{max}$ = 50\% (1307 Visual Tokens)}} \\
FoPru~\cite{jiang2024foprufocalpruningefficient} & 50.0\% & 1307 & 1485.38 & 87.23 & 62.13 & 65.53 & 78.31 & 47.90 & 63.25 & 65.45 & 243 & 17.78 \\
VisionZip~\cite{yang2024visionziplongerbetternecessary} & 50.0\% & 1307 & 1486.58 & 88.28 & 63.53 & 67.53 & 77.98 & 50.50 & 63.75 & 65.84 & 244 & 17.50 \\
PruMerge~\cite{shang2024llavaprumergeadaptivetokenreduction} & 50.0\% & 1307 & 1286.08 & 79.33 & 50.03 & 62.63 & 78.57 & 38.40 & 61.73 & 63.86 & 247 & 17.65 \\
HiRED~\cite{arif2024hiredattentionguidedtokendropping} & 50.0\% & 1307 & 1474.85 & 88.23 & 63.15 & 67.78 & 78.12 & \textbf{51.00} & 63.83 & 65.74 & 246 & 17.68 \\
\makecell[c]{HIVTP (Ours) \\ $(r{=}2, k{=}25, c{=}2)$} & 49.3\% & 1291 & \textbf{1487.49} & \textbf{88.34} & \textbf{63.80} & \textbf{67.97} & \textbf{78.92} & 48.40 & \textbf{64.03} & \textbf{66.06} & \textbf{242} & \textbf{17.85} \\
\midrule
\rowcolor{gray!15}
\multicolumn{13}{c}{\textit{$r_{max}$ = 40\% (1043 Visual Tokens)}} \\
FoPru~\cite{jiang2024foprufocalpruningefficient} & 40\% & 1043 & 1481.96 & 87.94 & 62.90 & 66.67 & 78.05 & \textbf{50.90} & 63.45 & 65.05 & 220 & 18.49 \\
VisionZip~\cite{yang2024visionziplongerbetternecessary} & 40.0\% & 1043 & 1484.58 & 87.79 & 63.23 & 66.58 & 78.17 & 49.80 & 63.58 & 65.65 & 237 & 18.15 \\
PruMerge~\cite{shang2024llavaprumergeadaptivetokenreduction} & 40.0\% & 1043 & 1270.50 & 79.12 & 49.32 & 61.34 & 78.16 & 37.70 & 61.32 & 63.86 & 248 & 16.97 \\
HiRED~\cite{arif2024hiredattentionguidedtokendropping} & 40.0\% & 1043 & 1469.12 & 88.00 & 63.53 & 67.44 & 77.95 & 50.70 & 63.66 & 65.68 & 216 & 18.79 \\
\makecell[c]{HIVTP (Ours) \\ $(r{=}2, k{=}15, c{=}2)$} & 39.4\% & 1031 & \textbf{1490.98} & \textbf{88.03} & \textbf{63.61} & \textbf{67.46} & \textbf{78.24} & 48.50 & \textbf{63.87} & \textbf{65.77} & \textbf{215} & \textbf{18.86} \\
\midrule
\rowcolor{gray!15}
\multicolumn{13}{c}{\textit{$r_{max}$ = 25\% (653 Visual Tokens)}} \\
FoPru~\cite{jiang2024foprufocalpruningefficient} & 25.0\% & 653 & 1474.91 & 86.56 & 62.23 & 65.81 & 78.02 & 48.10 & 61.87 & 65.48 & 200 & 18.86 \\
VisionZip~\cite{yang2024visionziplongerbetternecessary} & 25.0\% & 653 & 1459.96 & 87.31 & 62.10 & 65.46 & 78.47 & 48.20 & 62.89 & 65.61 & 233 & 18.84 \\
PruMerge~\cite{shang2024llavaprumergeadaptivetokenreduction} & 25.0\% & 653 & 1234.16 & 79.08 & 48.50 & 59.54 & 78.61 & 37.60 & 60.38 & 63.67 & 191 & \textbf{20.68} \\
HiRED~\cite{arif2024hiredattentionguidedtokendropping} & 25.0\% & 653 & 1467.24 & 87.36 & 62.28 & 65.98 & 78.38 & \textbf{48.60} & 63.24 & 65.67 & 201 & 19.47 \\
\makecell[c]{HIVTP (Ours) \\ $(r{=}2, k{=}14, c{=}3)$} & 24.8\% & 650 & \textbf{1493.89} & \textbf{87.69} & \textbf{62.79} & \textbf{66.52} & \textbf{78.73} & 47.44 & \textbf{63.65} & \textbf{65.87} & \textbf{190} & 19.54 \\
\bottomrule
\end{tabular}}
\endgroup
\caption{Accuracy and efficiency comparison of HIVTP with prior methods on LLaVA-Next-7B under different upper bounds of the visual token retaining ratio, $r_{\text{max}} \in \{75\%, 50\%, 40\%, 25\%\}$. The actual retaining ratio is denoted by $r_{\text{retain}}$. TTFT and Throughput are computed as averages across all benchmarks. The best results among all 
methods for each $r_{\text{max}}$ are highlighted in \textbf{bold}.}
\label{tab:main_llavanext}
\end{table*}

\subsection{Accuracy Analysis}
In Table~\ref{tab:main_llava15} 
and Table~\ref{tab:main_llavanext} 
we report the accuracy of our method across eight benchmarks 
on LLaVA-v1.5-7B and LLaVA-Next-7B, respectively, and compare it against prior methods~\cite{yang2024visionziplongerbetternecessary, jiang2024foprufocalpruningefficient, shang2024llavaprumergeadaptivetokenreduction, arif2024hiredattentionguidedtokendropping}.

\noindent\textbf{Accuracy on LLaVA-v1.5.} 
As shown in Table~\ref{tab:main_llava15}, compared to the original LLaVA-v1.5-7B, HIVTP achieves an average accuracy that is only 2.8\% lower across all tested values of $r_{\text{max}}$ and all benchmarks. It is worth noting that HIVTP achieves even higher accuracy than the original LLaVA-v1.5-7B on SQA across all tested values of $r_{\text{max}}$, and also surpasses it on MMB, OCR, and POPE under specific values of $r_{\text{max}}$. Moreover, compared to prior methods, HIVTP achieves higher accuracy on most benchmarks while using even fewer tokens across all values of $r_{\text{max}}$.

\noindent\textbf{Accuracy on LLaVA-Next.} 
As shown in Table~\ref{tab:main_llavanext}, compared with the original LLaVA-Next, HIVTP achieves higher accuracy on both the MME and SQA benchmarks under all values of $r_{\text{max}}$, with improvements of up to 1.7\% and 1.0\%, respectively. On the remaining benchmarks, HIVTP exhibits only a slight decrease of 0.8\% on average across all values of $r_{\text{max}}$. Besides, our method consistently outperforms prior methods on nearly all benchmarks across all values of $r_{\text{max}}$, while using even fewer visual tokens.

Overall, on both LLaVA-v1.5-7B and LLaVA-Next, HIVTP achieves higher accuracy across widely used benchmarks while using fewer visual tokens compared to prior methods, demonstrating its superiority over prior methods. 
Compared to the original LLaVA-v1.5-7B and LLaVA-Next models, implementing HIVTP can prune a large number of visual tokens without sacrificing much accuracy. In fact, it consistently achieves higher accuracy on SQA. We hypothesize that this is because the task is less dependent on visual information, 
and, by pruning less important visual tokens, performance can be improved to a certain extent.


\subsection{Efficiency Analysis}
To further quantify the advantage of our method in terms of inference efficiency, we adopt two widely used metrics: time-to-first-token (TTFT)~\cite{jiang2024foprufocalpruningefficient}, which measures the latency until the model generates the first token of the response, and token generation throughput (throughput)~\cite{arif2024hiredattentionguidedtokendropping}, which measures the number of tokens generated by the model per second. 
We compute the average results of HIVTP and prior methods across all benchmarks, and the results on LLaVA-v1.5-7B and LLaVA-Next are presented in Table~\ref{tab:main_llava15} and Table~\ref{tab:main_llavanext}, respectively.

HIVTP significantly improves the inference efficiency of both LLaVA-v1.5-7B and LLaVA-Next-7B. Specifically, compared to the original models, at the minimum tested $r_{\text{max}}$ of 25\%, HIVTP reduces TTFT by up to 50.0\% and 55.1\%, respectively, and increases throughput by up to 60.9\% and 47.3\%, respectively.
Additionally, HIVTP achieves higher inference efficiency than prior methods overall, mainly because it retains fewer visual tokens under the same $r_{\text{max}}$. Compared to VisionZip, HIVTP reduces TTFT by 26.5\% under $r_{\text{max}} = 40\%$ on LLaVA-v1.5-7B. Compared to PruMerge, HIVTP increases throughput by 11.1\% under $r_{\text{max}} = 40\%$ on LLaVA-Next-7B. On LLaVA-v1.5-7B, the average TTFT of HIVTP across different $r_{\text{max}}$ values is 4.0\,ms lower than that of FoPru, which has the lowest average TTFT among prior methods. The average throughput of HIVTP exceeds that of HiRED, the highest among prior methods, by 0.8 tokens/second. On LLaVA-Next-7B, HIVTP achieves an average TTFT that is 9.3\,ms lower than HiRED, the prior method with the lowest TTFT, and an average throughput that is 0.21 tokens/second higher than HiRED, which has the highest throughput among prior methods.

\section{Ablation}
In this study, we evaluate the influence of key components of HIVTP on model performance. The results are shown in Table~\ref{tab:ablation}. For fairness, all experiments are conducted on the LLaVA-v1.5-7B using the MME, POPE, and Ai2D benchmarks, with the upper bound of the visual token retaining ratio set to $r_{\text{max}} = 50\%$.

\noindent\textbf{Influence of Layers.} 
As shown in Table~\ref{tab:ablation}, using the middle layers of the vision encoder, $L_s = \{l_7, l_8, l_9, l_{10}\}$, to compute the importance scores of visual tokens yields better performance across all benchmarks compared to using later layers $L_s = \{l_{23}\}$~\cite{yang2024visionziplongerbetternecessary, jiang2024foprufocalpruningefficient, shang2024llavaprumergeadaptivetokenreduction, arif2024hiredattentionguidedtokendropping}, $L_s = \{l_{20}, l_{21}, l_{22}, l_{23}\}$, and earlier layers $L_s = \{l_1, l_2, l_3, l_4\}$. This result may be attributed to the fact that attention maps from middle layers are more effective at identifying fine-grained and object-level visual tokens, whereas attention maps from later layers and earlier layers can only identify high-level semantic tokens and struggle to estimate visual token importance, respectively.

\noindent\textbf{Influence of Global Region Size.}
We further evaluate the influence on how we divide the image into 
regions in the global retaining stage on model performance. 
As shown in Table~\ref{tab:ablation}, 
HIVTP (r=1, w/o region division) refers to retaining the top-25\% important tokens directly from all visual tokens in the global retaining stage without region division, and applying $2 \times 2$ windows in the local retaining stage. It performs worse than those that incorporate region division.
This may be because the retaining process in this case relies solely on visual importance scores without explicitly considering the spatial distribution of tokens in the original image, resulting in globally important visual tokens being unevenly distributed and thus degrading performance. However, dividing the image into 9 global regions (HIVTP (r=3)) performs worse than dividing it into 4 global regions (HIVTP (r=2)), possibly because excessive region division leads to too few visual tokens within each region, making it difficult to retain globally representative tokens with high importance scores.

\begin{table}[t]
\centering
\renewcommand{\arraystretch}{0.8}
{\fontsize{7pt}{8.5pt}\selectfont
\setlength{\tabcolsep}{0.9pt}
\begin{tabularx}{\linewidth}{@{}lccc@{}}
\toprule
\textbf{Choice} & \textbf{MME}~\cite{fu2024mmecomprehensiveevaluationbenchmark}$\uparrow$ & \textbf{POPE}~\cite{li2023evaluatingobjecthallucinationlarge}$\uparrow$ & \textbf{Ai2D}~\cite{kembhavi2016diagramworthdozenimages}$\uparrow$ \\
\midrule
\multicolumn{4}{l}{\textbf{Influence of Layers}} \\
\multicolumn{4}{l}{$(r{=}2, k{=}25, c{=}2, r_{max}=50\%)$} \\
\quad HIVTP($L_s = \{l_{23}\}$) & 1480.08 & 86.15 & 54.06 \\
\quad HIVTP($L_s = \{l_{20},l_{21},l_{22},l_{23}\}$) & 1481.49 & 87.02 & 54.76 \\
\quad \textbf{HIVTP($L_s = \{l_7,l_8,l_9,l_{10}\}$)} & \textbf{1495.37} & \textbf{87.12} & \textbf{54.99} \\
\quad HIVTP($L_s = \{l_{1},l_{2},l_{3},l_{4}\}$) & 1410.26 & 84.60 & 54.21 \\
\midrule
\multicolumn{4}{l}{\textbf{Influence of Region Division}} \\
\multicolumn{4}{l}{$(k{=}25, c{=}2, r_{max}=50\%, L_s = \{l_7,l_8,l_9,l_{10}\})$} \\
\quad HIVTP(r=1, w/o region division) & 1469.70 & 86.39 & 54.63 \\
\quad \textbf{HIVTP(r=2)} & \textbf{1495.37} & \textbf{87.12} & \textbf{54.99} \\
\quad HIVTP(r=3) & 1476.76 & 86.78 & 54.89 \\
\midrule
\multicolumn{4}{l}{\textbf{Influence of Window Division}} \\
\multicolumn{4}{l}{$(r{=}2, r_{max}=50\%, L_s = \{l_7,l_8,l_9,l_{10}\})$} \\
\quad HIVTP($k=50$, w/o local retaining) & 1453.66 & 86.92 & 54.84 \\
\quad \textbf{HIVTP($k=25$, c=2)} & \textbf{1495.37} & \textbf{87.12} & \textbf{54.99} \\
\quad HIVTP($k=39$, c=3) & 1473.86 & 87.04 & 54.93 \\
\bottomrule
\end{tabularx}
}
\caption{Ablation study on layer selection for computing visual token importance scores, as well as region division and window division in HIVTP.}
\label{tab:ablation}
\end{table}

\begin{figure}[t]
    \centering
    \includegraphics[width=0.9\linewidth]{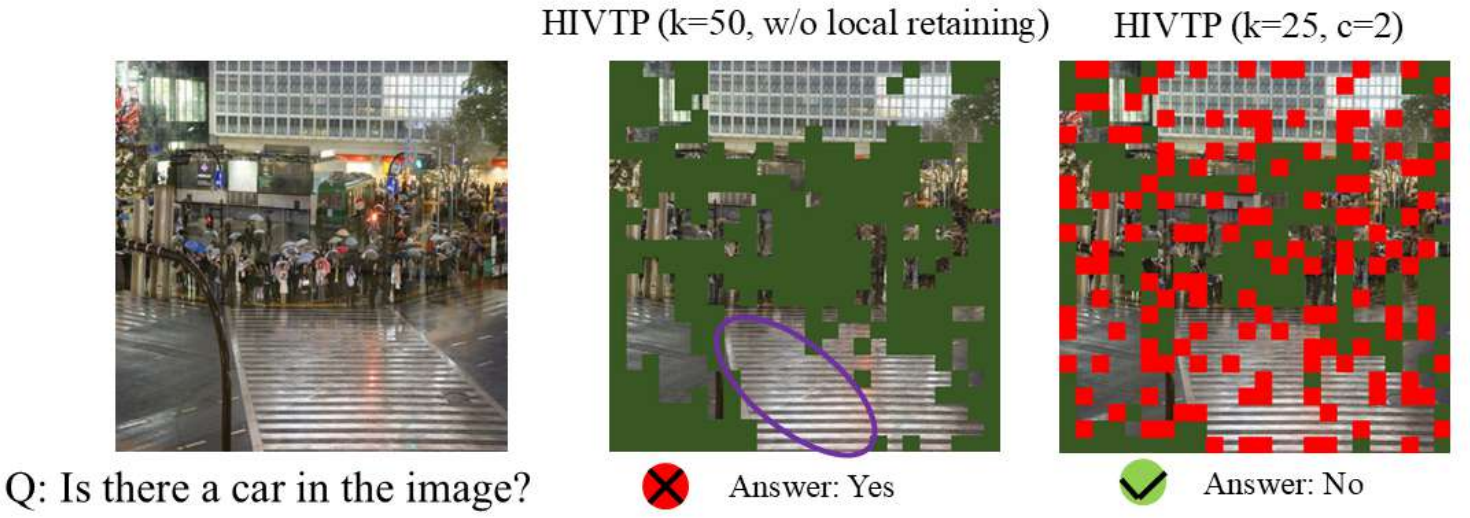}
    \caption{Comparison of HIVTP with and without the local retaining stage on a POPE example. HIVTP ($k=50$, w/o local retaining) denotes the variant without the local retaining stage, which hallucinates a car in the image. In contrast, HIVTP ($k=25$, c=2), which applies a window size of $2 \times 2$ in the local retaining stage, avoids hallucination. Green areas indicate the retained globally important visual tokens, while red areas indicate the retained locally important visual tokens. The region marked by the purple ellipse highlights a large contiguous area of pruned tokens.}
    \label{fig:hallucination}
\end{figure}

\noindent\textbf{Influence of Window Division.} In Table~\ref{tab:ablation}, HIVTP ($k=50$, w/o local retaining) refers to the hierarchical visual token pruning strategy that retains the top-50\% important tokens from each of the four regions in the global retaining stage only. 
HIVTP ($k=39$, c=3) and HIVTP ($k=25$, c=2) refer to retaining the top-39\% and top-25\% important tokens from each of the four regions in the global retaining stage and applying $3 \times 3$ and $2 \times 2$ windows in the local retaining stage, respectively. We set their $k$ to 39 and 25 to ensure that the upper bound of the visual token retaining ratio is $r_{\text{max}} = 50\%$.
As shown in Table~\ref{tab:ablation}, HIVTP with local retaining consistently outperforms the variant without local retaining across all benchmarks, demonstrating the benefit of retaining locally important fine-grained visual tokens for model accuracy. In particular, HIVTP with local retaining achieves better performance on the POPE benchmark, further indicating that local retaining helps mitigate hallucination in VLMs. We hypothesize that this improvement stems from the reduced visual uncertainty, as prior studies~\cite{ding2024hallupievaluatinghallucinationmultimodal, leng2023mitigatingobjecthallucinationslarge} have shown that visual uncertainty amplifies hallucination in VLMs. 
To empirically validate our hypothesis that local retaining can mitigate hallucination in VLMs to some extent, we visualize an example from the POPE benchmark to illustrate the difference between HIVTP ($k=50$, w/o local retaining) and HIVTP ($k=25$, c=2) in Figure~\ref{fig:hallucination}. 
Compared to HIVTP ($k=50$, w/o local retaining), HIVTP ($k=25$, c=2) promotes a more spatially uniform retention of visual tokens, effectively avoiding the dropping of large contiguous regions (as indicated by the purple ellipse) and thereby reducing visual uncertainty. This helps avoid hallucinating a car in the image. More examples from POPE are provided in Appendix~\ref{subsec:hallucination} .

\section{Conclusion and Limitations}
\label{sec:Methodology}
This paper proposes HIVTP, a training-free method to improve VLMs inference efficiency via hierarchical visual token pruning using a novel middle-layer-based importance score. We leverage attention maps from selected middle layers of the vision encoder to estimate the importance of visual tokens, as these layers better capture fine-grained and object-level attention. We then propose a hierarchical visual token pruning strategy. In the global retaining stage, we perform region division and retain tokens with higher importance scores in each region to obtain globally important visual tokens. In the local retaining stage, we perform window division and retain the token with the highest importance score in each window to obtain locally important visual tokens. Extensive evaluations on two LLaVA models using widely adopted benchmarks demonstrate that HIVTP can significantly improve inference efficiency with minimal accuracy loss, and even improve accuracy on certain benchmarks.

Despite the strong performance demonstrated by HIVTP, it still has limitations. Specifically, the token retaining ratio in our method is manually set, whereas the optimal ratio may vary across different visual tasks. Future work could explore how to dynamically adjust the token retaining ratio in a task-wise manner.

{
    \small
    \bibliographystyle{ieeenat_fullname}
    \bibliography{main}

\begin{thebibliography}{44}
\providecommand{\natexlab}[1]{#1}
\providecommand{\url}[1]{\texttt{#1}}
\expandafter\ifx\csname urlstyle\endcsname\relax
  \providecommand{\doi}[1]{doi: #1}\else
  \providecommand{\doi}{doi: \begingroup \urlstyle{rm}\Url}\fi

\bibitem[Ainslie et~al.(2023)Ainslie, Lee-Thorp, de~Jong, Zemlyanskiy, Lebrón, and Sanghai]{ainslie2023gqatraininggeneralizedmultiquery}
Joshua Ainslie, James Lee-Thorp, Michiel de Jong, Yury Zemlyanskiy, Federico Lebrón, and Sumit Sanghai.
\newblock {GQA}: Training generalized multi-query transformer models from multi-head checkpoints, 2023.

\bibitem[Arif et~al.(2024)Arif, Yoon, Nikolopoulos, Vandierendonck, John, and Ji]{arif2024hiredattentionguidedtokendropping}
Kazi Hasan~Ibn Arif, JinYi Yoon, Dimitrios~S. Nikolopoulos, Hans Vandierendonck, Deepu John, and Bo Ji.
\newblock {HiRED}: Attention-guided token dropping for efficient inference of high-resolution vision-language models, 2024.

\bibitem[Bai et~al.(2023)Bai, Bai, Yang, Wang, Tan, Wang, Lin, Zhou, and Zhou]{bai2023qwenvlversatilevisionlanguagemodel}
Jinze Bai, Shuai Bai, Shusheng Yang, Shijie Wang, Sinan Tan, Peng Wang, Junyang Lin, Chang Zhou, and Jingren Zhou.
\newblock {Qwen-VL}: A versatile vision-language model for understanding, localization, text reading, and beyond, 2023.

\bibitem[Cha et~al.(2024)Cha, Kang, Mun, and Roh]{cha2024honeybee}
Junbum Cha, Wooyoung Kang, Jonghwan Mun, and Byungseok Roh.
\newblock Honeybee: Locality-enhanced projector for multimodal {LLM}.
\newblock In \emph{Proceedings of the IEEE/CVF Conference on Computer Vision and Pattern Recognition}, pages 13817--13827, 2024.

\bibitem[Chen et~al.(2024{\natexlab{a}})Chen, Li, Dong, Zhang, He, Wang, Zhao, and Lin]{chen2024sharegpt4v}
Lin Chen, Jinsong Li, Xiaoyi Dong, Pan Zhang, Conghui He, Jiaqi Wang, Feng Zhao, and Dahua Lin.
\newblock {ShareGPT4V}: Improving large multi-modal models with better captions.
\newblock In \emph{European Conference on Computer Vision}, pages 370--387. Springer, 2024{\natexlab{a}}.

\bibitem[Chen et~al.(2024{\natexlab{b}})Chen, Wu, Wang, Su, Chen, Xing, Zhong, Zhang, Zhu, Lu, et~al.]{chen2024internvl}
Zhe Chen, Jiannan Wu, Wenhai Wang, Weijie Su, Guo Chen, Sen Xing, Muyan Zhong, Qinglong Zhang, Xizhou Zhu, Lewei Lu, et~al.
\newblock {InternVL}: Scaling up vision foundation models and aligning for generic visual-linguistic tasks.
\newblock In \emph{Proceedings of the IEEE/CVF conference on computer vision and pattern recognition}, pages 24185--24198, 2024{\natexlab{b}}.

\bibitem[Chiang et~al.(2023)Chiang, Li, Lin, Sheng, Wu, Zhang, Zheng, Zhuang, Zhuang, Gonzalez, Stoica, and Xing]{vicuna2023}
Wei-Lin Chiang, Zhuohan Li, Zi Lin, Ying Sheng, Zhanghao Wu, Hao Zhang, Lianmin Zheng, Siyuan Zhuang, Yonghao Zhuang, Joseph~E. Gonzalez, Ion Stoica, and Eric~P. Xing.
\newblock Vicuna: An open-source chatbot impressing {GPT-4} with 90\%* {ChatGPT }quality, 2023.

\bibitem[Comanici et~al.(2025)Comanici, Bieber, Schaekermann, Pasupat, Sachdeva, Dhillon, Blistein, Ram, Zhang, Rosen, et~al.]{comanici2025gemini}
Gheorghe Comanici, Eric Bieber, Mike Schaekermann, Ice Pasupat, Noveen Sachdeva, Inderjit Dhillon, Marcel Blistein, Ori Ram, Dan Zhang, Evan Rosen, et~al.
\newblock Gemini 2.5: Pushing the frontier with advanced reasoning, multimodality, long context, and next generation agentic capabilities.
\newblock \emph{arXiv preprint arXiv:2507.06261}, 2025.

\bibitem[Ding et~al.(2024)Ding, Wu, Kuang, Ma, Cao, Cai, Chen, Chen, and Huang]{ding2024hallupievaluatinghallucinationmultimodal}
Peng Ding, Jingyu Wu, Jun Kuang, Dan Ma, Xuezhi Cao, Xunliang Cai, Shi Chen, Jiajun Chen, and Shujian Huang.
\newblock {Hallu-PI}: Evaluating hallucination in multi-modal large language models within perturbed inputs, 2024.

\bibitem[Fu et~al.(2024)Fu, Chen, Shen, Qin, Zhang, Lin, Yang, Zheng, Li, Sun, Wu, and Ji]{fu2024mmecomprehensiveevaluationbenchmark}
Chaoyou Fu, Peixian Chen, Yunhang Shen, Yulei Qin, Mengdan Zhang, Xu Lin, Jinrui Yang, Xiawu Zheng, Ke Li, Xing Sun, Yunsheng Wu, and Rongrong Ji.
\newblock {MME}: A comprehensive evaluation benchmark for multimodal large language models, 2024.

\bibitem[Jiang et~al.(2024)Jiang, Huang, Liu, Zeng, Li, Cheng, and Xu]{jiang2024foprufocalpruningefficient}
Lei Jiang, Weizhe Huang, Tongxuan Liu, Yuting Zeng, Jing Li, Lechao Cheng, and Xiaohua Xu.
\newblock {FoPru}: Focal pruning for efficient large vision-language models, 2024.

\bibitem[Kembhavi et~al.(2016)Kembhavi, Salvato, Kolve, Seo, Hajishirzi, and Farhadi]{kembhavi2016diagramworthdozenimages}
Aniruddha Kembhavi, Mike Salvato, Eric Kolve, Minjoon Seo, Hannaneh Hajishirzi, and Ali Farhadi.
\newblock A diagram is worth a dozen images, 2016.

\bibitem[Leng et~al.(2023)Leng, Zhang, Chen, Li, Lu, Miao, and Bing]{leng2023mitigatingobjecthallucinationslarge}
Sicong Leng, Hang Zhang, Guanzheng Chen, Xin Li, Shijian Lu, Chunyan Miao, and Lidong Bing.
\newblock Mitigating object hallucinations in large vision-language models through visual contrastive decoding, 2023.

\bibitem[Li et~al.(2025{\natexlab{a}})Li, Yang, and Lu]{li2025todrevisualtokenpruning}
Duo Li, Zuhao Yang, and Shijian Lu.
\newblock {ToDRE}: Visual token pruning via diversity and task awareness for efficient large vision-language models, 2025{\natexlab{a}}.

\bibitem[Li et~al.(2023{\natexlab{a}})Li, Li, Savarese, and Hoi]{li2023blip}
Junnan Li, Dongxu Li, Silvio Savarese, and Steven Hoi.
\newblock {BLIP-2}: Bootstrapping language-image pre-training with frozen image encoders and large language models.
\newblock In \emph{International conference on machine learning}, pages 19730--19742. PMLR, 2023{\natexlab{a}}.

\bibitem[Li et~al.(2025{\natexlab{b}})Li, Goyal, Semedo, and Kolter]{li2025inferenceoptimalvlmsneed}
Kevin~Y. Li, Sachin Goyal, Joao~D. Semedo, and J.~Zico Kolter.
\newblock Inference optimal {VLMs} need fewer visual tokens and more parameters, 2025{\natexlab{b}}.

\bibitem[Li et~al.(2025{\natexlab{c}})Li, Yuan, Liu, Tang, Wang, Qin, Zhu, and Zhang]{li2025tokenpacker}
Wentong Li, Yuqian Yuan, Jian Liu, Dongqi Tang, Song Wang, Jie Qin, Jianke Zhu, and Lei Zhang.
\newblock {TokenPacker}: Efficient visual projector for multimodal {LLM}.
\newblock \emph{International Journal of Computer Vision}, pages 1--19, 2025{\natexlab{c}}.

\bibitem[Li et~al.(2023{\natexlab{b}})Li, Du, Zhou, Wang, Zhao, and Wen]{li2023evaluatingobjecthallucinationlarge}
Yifan Li, Yifan Du, Kun Zhou, Jinpeng Wang, Wayne~Xin Zhao, and Ji-Rong Wen.
\newblock Evaluating object hallucination in large vision-language models, 2023{\natexlab{b}}.

\bibitem[Li et~al.(2025{\natexlab{d}})Li, Bao, Ye, Tan, Chen, Liu, and Kong]{li2025windowtokenconcatenationefficient}
Yifan Li, Wentao Bao, Botao Ye, Zhen Tan, Tianlong Chen, Huan Liu, and Yu Kong.
\newblock Window token concatenation for efficient visual large language models, 2025{\natexlab{d}}.

\bibitem[Lin et~al.(2024)Lin, Yin, Ping, Lu, Molchanov, Tao, Mao, Kautz, Shoeybi, and Han]{lin2024vilapretrainingvisuallanguage}
Ji Lin, Hongxu Yin, Wei Ping, Yao Lu, Pavlo Molchanov, Andrew Tao, Huizi Mao, Jan Kautz, Mohammad Shoeybi, and Song Han.
\newblock {VILA}: On pre-training for visual language models, 2024.

\bibitem[Liu et~al.(2023)Liu, Li, Wu, and Lee]{liu2023visualinstructiontuning}
Haotian Liu, Chunyuan Li, Qingyang Wu, and Yong~Jae Lee.
\newblock Visual instruction tuning, 2023.

\bibitem[Liu et~al.(2024{\natexlab{a}})Liu, Li, Li, and Lee]{liu2024improvedbaselinesvisualinstruction}
Haotian Liu, Chunyuan Li, Yuheng Li, and Yong~Jae Lee.
\newblock Improved baselines with visual instruction tuning, 2024{\natexlab{a}}.

\bibitem[Liu et~al.(2024{\natexlab{b}})Liu, Li, Li, Li, Zhang, Shen, and Lee]{liu2024llavanext}
Haotian Liu, Chunyuan Li, Yuheng Li, Bo Li, Yuanhan Zhang, Sheng Shen, and Yong~Jae Lee.
\newblock {LLaVA-NeXT}: Improved reasoning, {OCR}, and world knowledge, 2024{\natexlab{b}}.

\bibitem[Liu et~al.(2024{\natexlab{c}})Liu, Duan, Zhang, Li, Zhang, Zhao, Yuan, Wang, He, Liu, Chen, and Lin]{liu2024mmbenchmultimodalmodelallaround}
Yuan Liu, Haodong Duan, Yuanhan Zhang, Bo Li, Songyang Zhang, Wangbo Zhao, Yike Yuan, Jiaqi Wang, Conghui He, Ziwei Liu, Kai Chen, and Dahua Lin.
\newblock {MMBench}: Is your multi-modal model an all-around player?, 2024{\natexlab{c}}.

\bibitem[Liu et~al.(2024{\natexlab{d}})Liu, Li, Huang, Yang, Yu, Li, Yin, Liu, Jin, and Bai]{Liu_2024}
Yuliang Liu, Zhang Li, Mingxin Huang, Biao Yang, Wenwen Yu, Chunyuan Li, Xu-Cheng Yin, Cheng-Lin Liu, Lianwen Jin, and Xiang Bai.
\newblock {OCRBench}: on the hidden mystery of {OCR} in large multimodal models.
\newblock \emph{Science China Information Sciences}, 67\penalty0 (12), 2024{\natexlab{d}}.

\bibitem[Lu et~al.(2024)Lu, Liu, Zhang, Wang, Dong, Liu, Sun, Ren, Li, Yang, et~al.]{lu2024deepseek}
Haoyu Lu, Wen Liu, Bo Zhang, Bingxuan Wang, Kai Dong, Bo Liu, Jingxiang Sun, Tongzheng Ren, Zhuoshu Li, Hao Yang, et~al.
\newblock {DeepSeek-VL}: towards real-world vision-language understanding.
\newblock \emph{arXiv preprint arXiv:2403.05525}, 2024.

\bibitem[Lu et~al.(2022)Lu, Mishra, Xia, Qiu, Chang, Zhu, Tafjord, Clark, and Kalyan]{lu2022learnexplainmultimodalreasoning}
Pan Lu, Swaroop Mishra, Tony Xia, Liang Qiu, Kai-Wei Chang, Song-Chun Zhu, Oyvind Tafjord, Peter Clark, and Ashwin Kalyan.
\newblock Learn to explain: Multimodal reasoning via thought chains for science question answering, 2022.

\bibitem[OpenAI(August 7 2025)]{openai2025gpt5}
OpenAI.
\newblock Introducing {GPT-5}.
\newblock \url{https://openai.com/index/introducing-gpt-5/}, August 7 2025.
\newblock Accessed: 2025-09-17.

\bibitem[O'Shea and Nash(2015)]{oshea2015introductionconvolutionalneuralnetworks}
Keiron O'Shea and Ryan Nash.
\newblock An introduction to convolutional neural networks, 2015.

\bibitem[Radford et~al.(2021)Radford, Kim, Hallacy, Ramesh, Goh, Agarwal, Sastry, Askell, Mishkin, Clark, Krueger, and Sutskever]{radford2021learningtransferablevisualmodels}
Alec Radford, Jong~Wook Kim, Chris Hallacy, Aditya Ramesh, Gabriel Goh, Sandhini Agarwal, Girish Sastry, Amanda Askell, Pamela Mishkin, Jack Clark, Gretchen Krueger, and Ilya Sutskever.
\newblock Learning transferable visual models from natural language supervision, 2021.

\bibitem[Shang et~al.(2024)Shang, Cai, Xu, Lee, and Yan]{shang2024llavaprumergeadaptivetokenreduction}
Yuzhang Shang, Mu Cai, Bingxin Xu, Yong~Jae Lee, and Yan Yan.
\newblock {LLaVA-PruMerge}: Adaptive token reduction for efficient large multimodal models, 2024.

\bibitem[Singh et~al.(2019)Singh, Natarajan, Shah, Jiang, Chen, Batra, Parikh, and Rohrbach]{singh2019vqamodelsread}
Amanpreet Singh, Vivek Natarajan, Meet Shah, Yu Jiang, Xinlei Chen, Dhruv Batra, Devi Parikh, and Marcus Rohrbach.
\newblock Towards {VQA} models that can read, 2019.

\bibitem[Team et~al.(2024)Team, Georgiev, Lei, Burnell, Bai, Gulati, Tanzer, Vincent, Pan, Wang, et~al.]{team2024gemini}
Gemini Team, Petko Georgiev, Ving~Ian Lei, Ryan Burnell, Libin Bai, Anmol Gulati, Garrett Tanzer, Damien Vincent, Zhufeng Pan, Shibo Wang, et~al.
\newblock Gemini 1.5: Unlocking multimodal understanding across millions of tokens of context.
\newblock \emph{arXiv preprint arXiv:2403.05530}, 2024.

\bibitem[Vaswani et~al.(2023)Vaswani, Shazeer, Parmar, Uszkoreit, Jones, Gomez, Kaiser, and Polosukhin]{vaswani2023attentionneed}
Ashish Vaswani, Noam Shazeer, Niki Parmar, Jakob Uszkoreit, Llion Jones, Aidan~N. Gomez, Lukasz Kaiser, and Illia Polosukhin.
\newblock Attention is all you need, 2023.

\bibitem[Wang et~al.(2024)Wang, Lv, Yu, Hong, Qi, Wang, Ji, Yang, Zhao, XiXuan, et~al.]{wang2024cogvlm}
Weihan Wang, Qingsong Lv, Wenmeng Yu, Wenyi Hong, Ji Qi, Yan Wang, Junhui Ji, Zhuoyi Yang, Lei Zhao, Song XiXuan, et~al.
\newblock Cog{VLM}: Visual expert for pretrained language models.
\newblock \emph{Advances in Neural Information Processing Systems}, 37:\penalty0 121475--121499, 2024.

\bibitem[Wen et~al.(2024)Wen, Cao, Fu, Mehta, and Najibi]{wen2024efficientvisionlanguagemodelssummarizing}
Yuxin Wen, Qingqing Cao, Qichen Fu, Sachin Mehta, and Mahyar Najibi.
\newblock Efficient vision-language models by summarizing visual tokens into compact registers, 2024.

\bibitem[Wen et~al.(2025)Wen, Gao, Li, He, and Zhang]{wen2025tokenpruningmultimodallarge}
Zichen Wen, Yifeng Gao, Weijia Li, Conghui He, and Linfeng Zhang.
\newblock Token pruning in multimodal large language models: Are we solving the right problem?, 2025.

\bibitem[Yang et~al.(2024)Yang, Chen, Tian, Wang, Li, Yu, and Jia]{yang2024visionziplongerbetternecessary}
Senqiao Yang, Yukang Chen, Zhuotao Tian, Chengyao Wang, Jingyao Li, Bei Yu, and Jiaya Jia.
\newblock {VisionZip}: Longer is better but not necessary in vision language models, 2024.

\bibitem[Ye et~al.(2024)Ye, Gan, Ge, Zhang, and Tang]{ye2024atpllavaadaptivetokenpruning}
Xubing Ye, Yukang Gan, Yixiao Ge, Xiao-Ping Zhang, and Yansong Tang.
\newblock {ATP-LLaVA}: Adaptive token pruning for large vision language models, 2024.

\bibitem[Zeng et~al.(2024)Zeng, Yang, Yang, Yang, and Lin]{10.1145/3664647.3681712}
Jingjie Zeng, Zhihao Yang, Qi Yang, Liang Yang, and Hongfei Lin.
\newblock Peeling back the layers: Interpreting the storytelling of {ViT}.
\newblock In \emph{Proceedings of the 32nd ACM International Conference on Multimedia}, pages 7298--7306, 2024.

\bibitem[Zhang et~al.(2025{\natexlab{a}})Zhang, Lyu, He, Ao, and Lin]{zhang2025adaptivevisualtokenpruning}
Hao Zhang, Mengsi Lyu, Chenrui He, Yulong Ao, and Yonghua Lin.
\newblock Towards adaptive visual token pruning for large multimodal models, 2025{\natexlab{a}}.

\bibitem[Zhang et~al.(2025{\natexlab{b}})Zhang, Cheng, Lu, Zhang, Zhuo, Cao, Guo, She, and Zhang]{zhang2025textvisualattentionexploitingvisual}
Qizhe Zhang, Aosong Cheng, Ming Lu, Renrui Zhang, Zhiyong Zhuo, Jiajun Cao, Shaobo Guo, Qi She, and Shanghang Zhang.
\newblock Beyond text-visual attention: Exploiting visual cues for effective token pruning in {VLMs}, 2025{\natexlab{b}}.

\bibitem[Zhang et~al.(2025{\natexlab{c}})Zhang, Fan, Ma, Zheng, Huang, Cheng, Gudovskiy, Okuno, Nakata, Keutzer, and Zhang]{zhang2025sparsevlmvisualtokensparsification}
Yuan Zhang, Chun-Kai Fan, Junpeng Ma, Wenzhao Zheng, Tao Huang, Kuan Cheng, Denis Gudovskiy, Tomoyuki Okuno, Yohei Nakata, Kurt Keutzer, and Shanghang Zhang.
\newblock {SparseVLM}: Visual token sparsification for efficient vision-language model inference, 2025{\natexlab{c}}.

\bibitem[Zhu et~al.(2023)Zhu, Chen, Shen, Li, and Elhoseiny]{zhu2023minigpt}
Deyao Zhu, Jun Chen, Xiaoqian Shen, Xiang Li, and Mohamed Elhoseiny.
\newblock {MiniGPT-4}: Enhancing vision-language understanding with advanced large language models.
\newblock \emph{arXiv preprint arXiv:2304.10592}, 2023.

\end{thebibliography}
}

\appendix
\clearpage
\setcounter{page}{1} 
\renewcommand{\thesection}{\Alph{section}} 
\setcounter{section}{0} 
\renewcommand{\thesubsection}{\Alph{section}.\arabic{subsection}} 
\maketitlesupplementary

\section{Appendix}
\label{sec:Appendix}
\subsection{Algorithm}
\label{subsec:alg}
The algorithmic details of the visual token importance scores and the hierarchical visual token pruning strategy are shown in Algorithm~\ref{alg:vis_token_importance} and Algorithm~\ref{alg:hierarchical}, respectively.

\subsection{Visualization of Attention Heatmap}
\label{subsec:heatmap}
\begin{figure*}[t]
    \centering
    \includegraphics[width=1\linewidth]{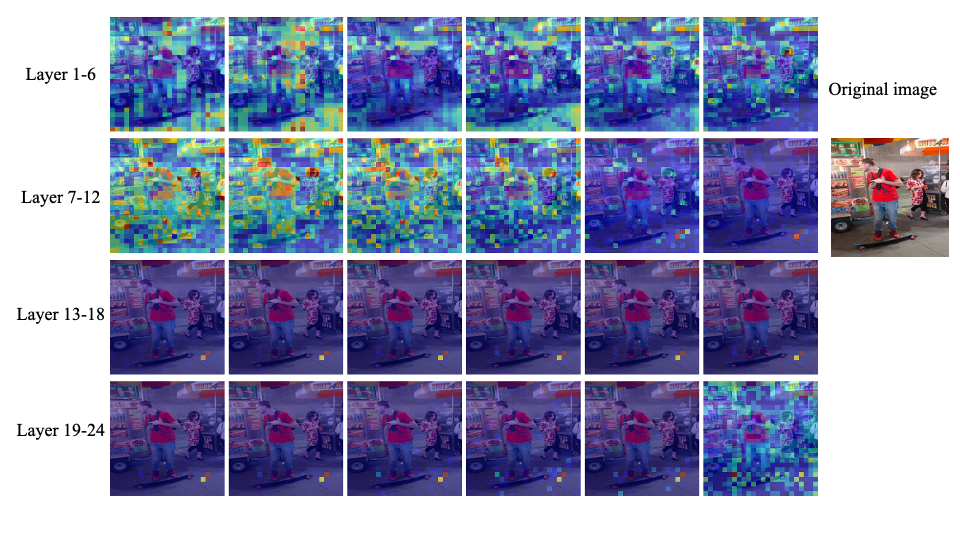}
    \caption{Attention heatmaps at different layers of the vision encoder in LLaVA-v1.5-7B for one example from the MME benchmark.}
    \label{fig:heatmap1}
\end{figure*}

\begin{figure*}[t]
    \centering
    \includegraphics[width=1\linewidth]{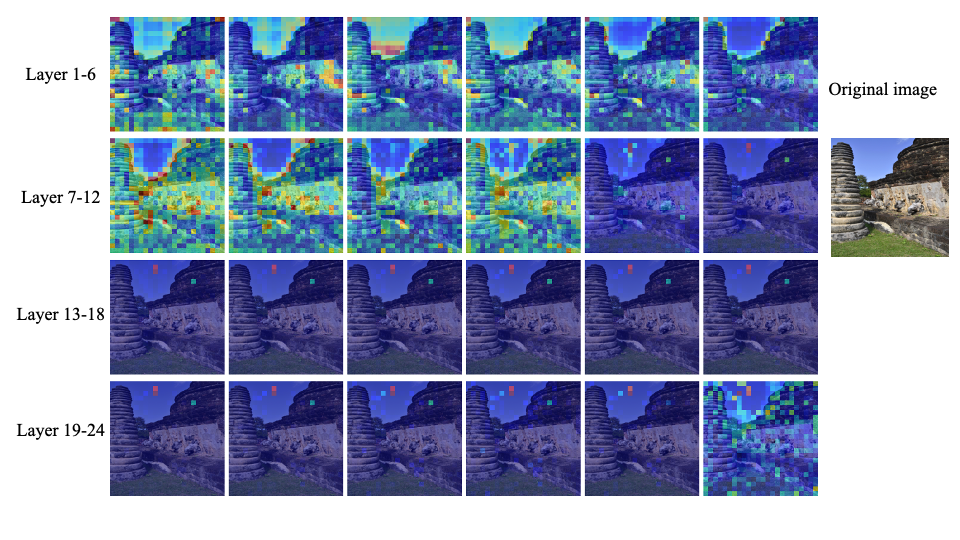}
    \caption{Attention heatmaps at different layers of the vision encoder in LLaVA-v1.5-7B for one example from the MME benchmark.}
    \label{fig:heatmap2}
\end{figure*}

\begin{figure*}[t]
    \centering
    \includegraphics[width=1\linewidth]{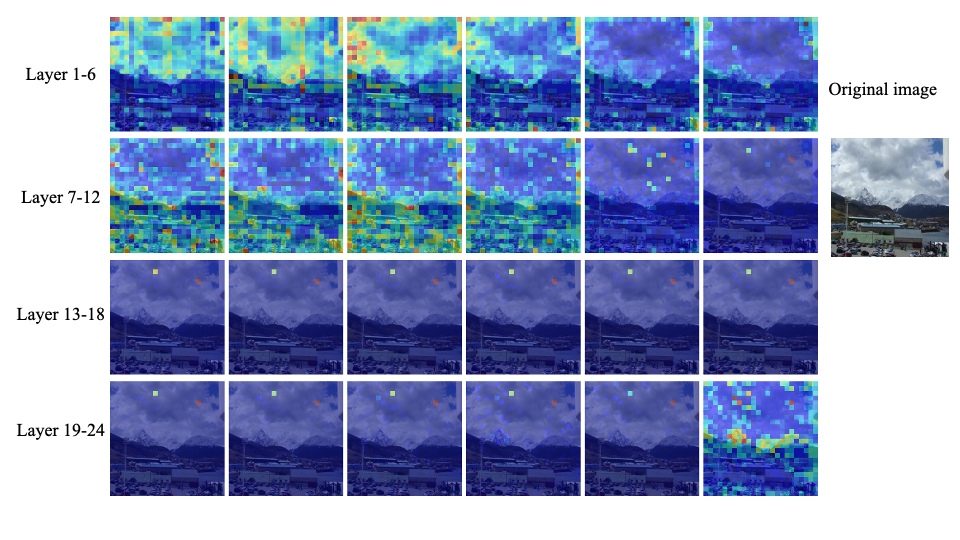}
    \caption{Attention heatmaps at different layers of the vision encoder in LLaVA-v1.5-7B for one example from the MME benchmark.}
    \label{fig:heatmap3}
\end{figure*}

\begin{figure*}[t]
    \centering
    \includegraphics[width=1\linewidth]{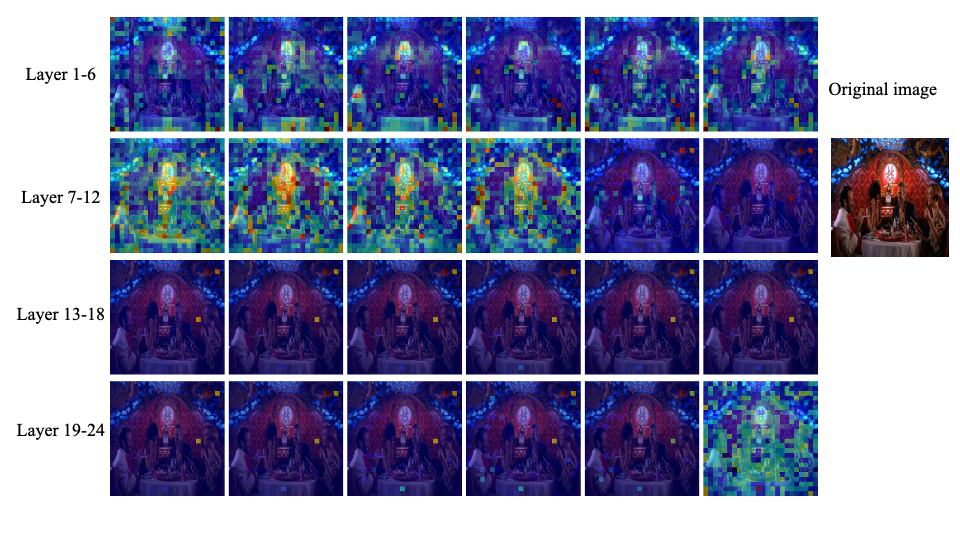}
    \caption{Attention heatmaps at different layers of the vision encoder in LLaVA-v1.5-7B for one example from the MME benchmark.}
    \label{fig:heatmap4}
\end{figure*}

\begin{figure*}[t]
    \centering
    \includegraphics[width=1\linewidth]{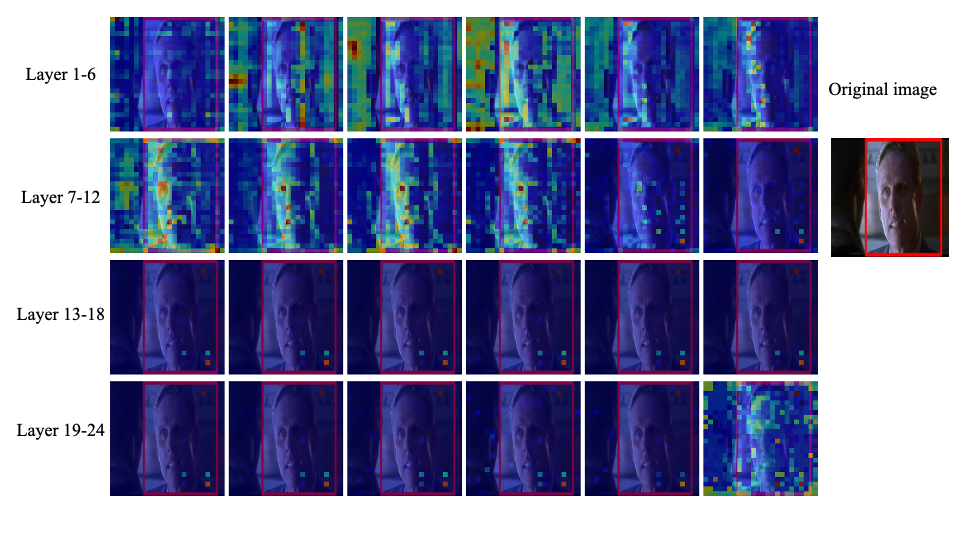}
    \caption{Attention heatmaps at different layers of the vision encoder in LLaVA-v1.5-7B for one example from the MME benchmark.}
    \label{fig:heatmap5}
\end{figure*}

\begin{figure*}[t]
    \centering
    \includegraphics[width=1\linewidth]{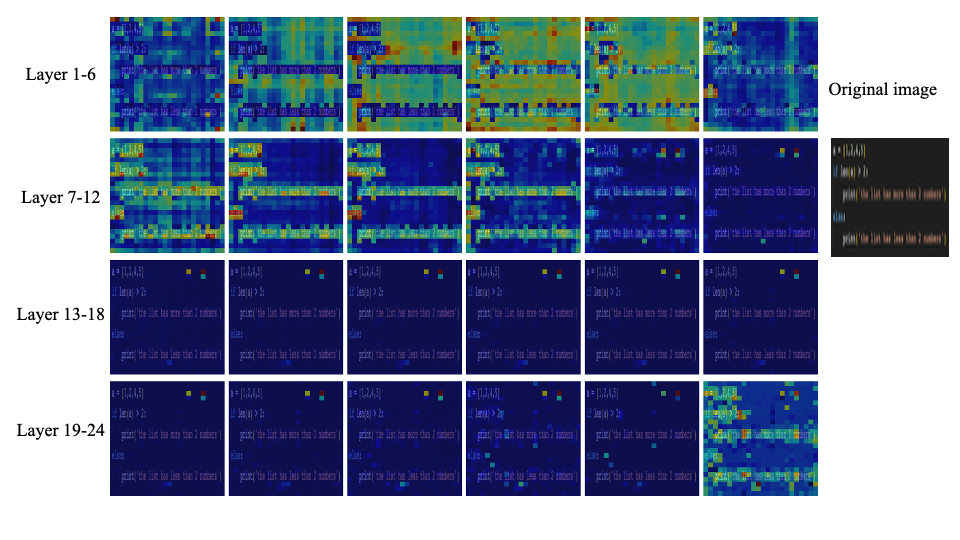}
    \caption{Attention heatmaps at different layers of the vision encoder in LLaVA-v1.5-7B for one example from the MME benchmark.}
    \label{fig:heatmap6}
\end{figure*}

To further demonstrate the behavioral characteristics of the vision encoder across different layer depths and to 
motivate the selection of the middle-layer index set $L_s = \{l_7, l_8, l_9, l_{10}\}$ to compute visual token importance scores in our method, we present additional examples from the MME benchmark showing the attention heatmaps at different layers of the vision encoder in LLaVA-v1.5-7B.
As shown in Figure~\ref{fig:heatmap1} to Figure~\ref{fig:heatmap6}, from layer 7 to layer 10, the visual tokens corresponding to the main objects in the image exhibit higher brightness, indicating higher attention weights.

\subsection{Visualization of POPE Examples}
\label{subsec:hallucination}
\begin{figure*}[t]
    \centering
    \includegraphics[width=1\linewidth]{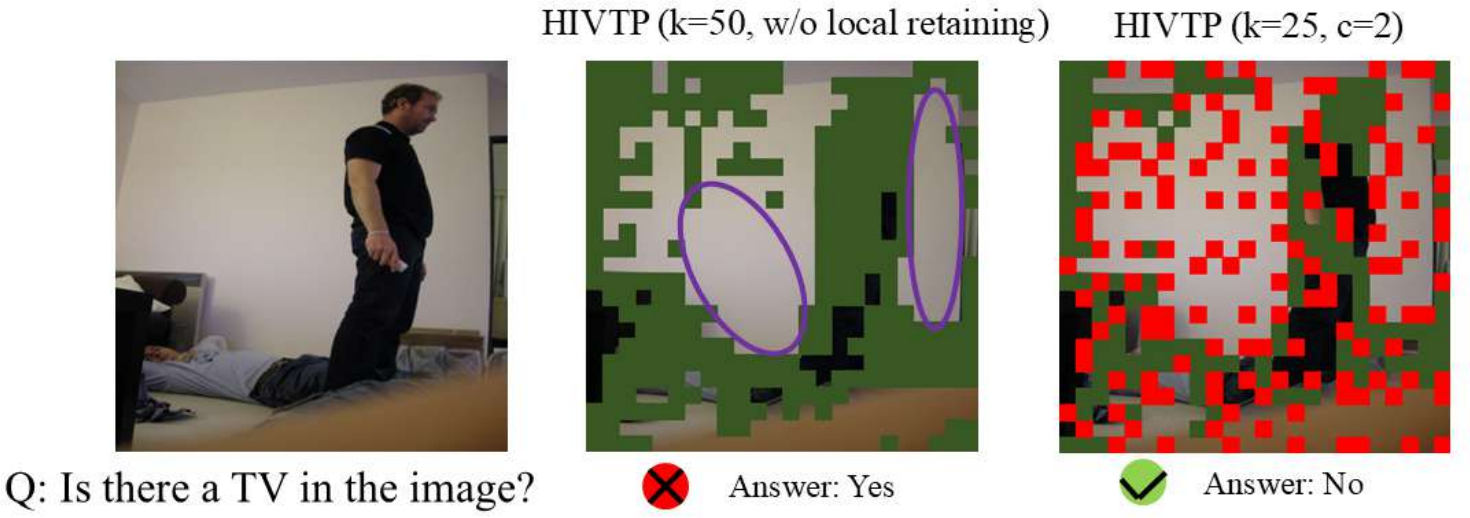}
    \caption{Comparison of HIVTP with and without the local retaining stage on a POPE example. HIVTP ($k=50$, w/o local retaining) denotes the variant without the local retaining stage; HIVTP ($k=25$, c=2) applies a window size of $2 \times 2$ in the local retaining stage.}
    \label{fig:hallucination1}
\end{figure*}

\begin{figure*}[t]
    \centering
    \includegraphics[width=1\linewidth]{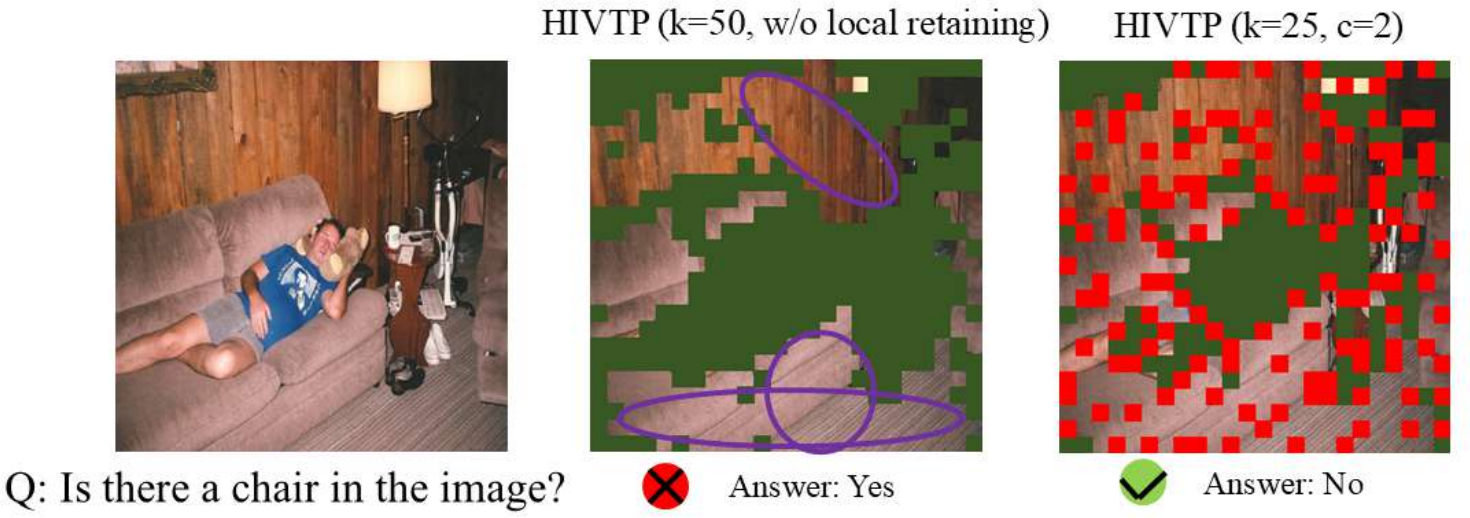}
    \caption{Comparison of HIVTP with and without the local retaining stage on a POPE example. HIVTP ($k=50$, w/o local retaining) denotes the variant without the local retaining stage; HIVTP ($k=25$, c=2) applies a window size of $2 \times 2$ in the local retaining stage.}
    \label{fig:hallucination2}
\end{figure*}

\begin{figure*}[t]
    \centering
    \includegraphics[width=1\linewidth]{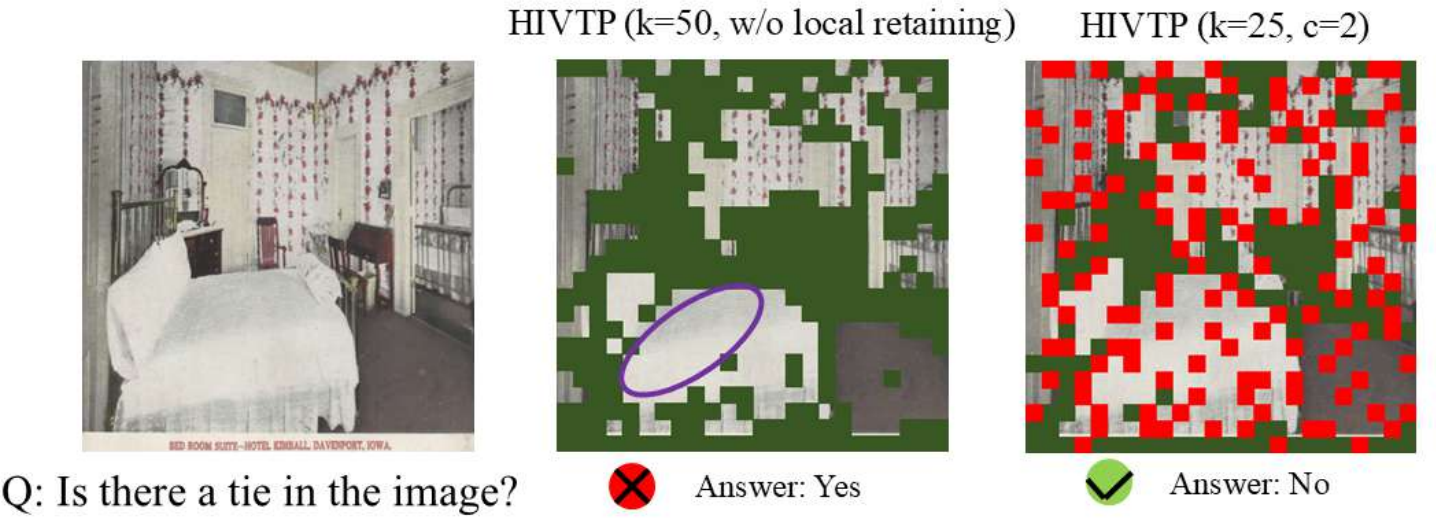}
    \caption{Comparison of HIVTP with and without the local retaining stage on a POPE example. HIVTP ($k=50$, w/o local retaining) denotes the variant without the local retaining stage; HIVTP ($k=25$, c=2) applies a window size of $2 \times 2$ in the local retaining stage.}
    \label{fig:hallucination3}
\end{figure*}

\begin{figure*}[t]
    \centering
    \includegraphics[width=1\linewidth]{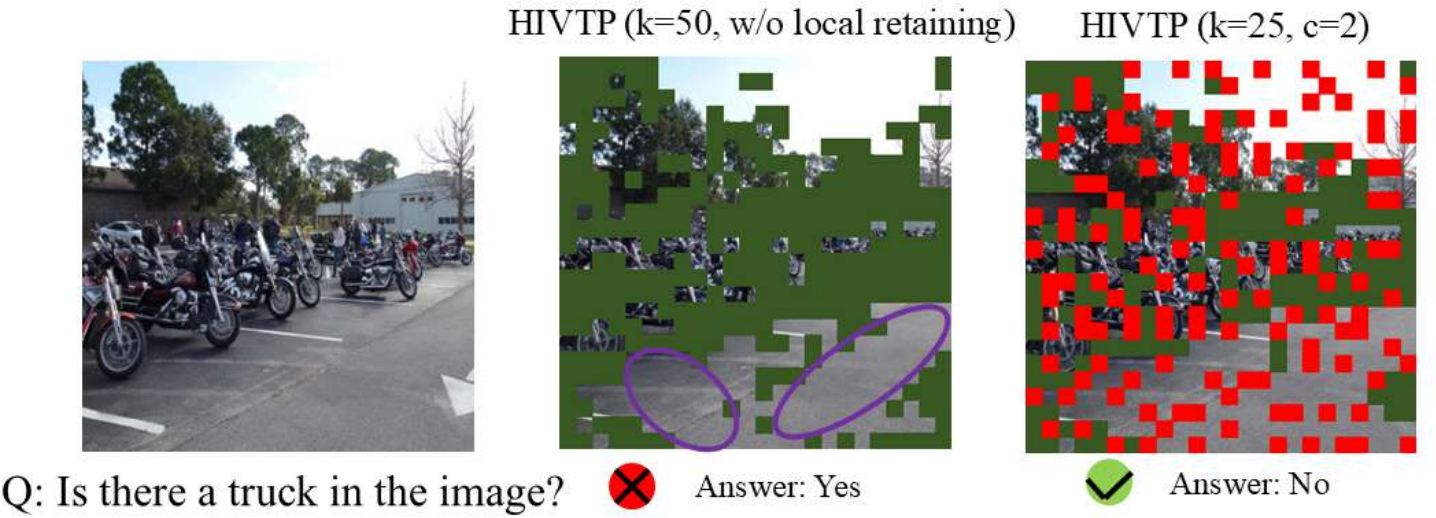}
    \caption{Comparison of HIVTP with and without the local retaining stage on a POPE example. HIVTP ($k=50$, w/o local retaining) denotes the variant without the local retaining stage; HIVTP ($k=25$, c=2) applies a window size of $2 \times 2$ in the local retaining stage.}
    \label{fig:hallucination4}
\end{figure*}

In the main text, we hypothesize that our local retaining in HIVTP can mitigate hallucination in VLMs due to reduced visual uncertainty. To validate this hypothesis, we visualize more examples from the POPE benchmark to illustrate the difference between HIVTP ($k=50$, w/o local retaining) and HIVTP ($k=25$, $c=2$). As shown in Figure~\ref{fig:hallucination1} to Figure~\ref{fig:hallucination4}, HIVTP ($k=50$, w/o local retaining) causes the dropping of large contiguous regions (as indicated by the purple ellipse), which increases visual uncertainty and results in hallucination across all these examples. In contrast, HIVTP ($k=25$, $c=2$), due to the use of local retaining, prevents the dropping of large contiguous regions and avoids hallucination. This visualization result supports our hypothesis that local retaining can mitigate hallucination by reducing visual uncertainty.

\begin{algorithm}[h]
\footnotesize
\SetNlSty{textbf}{\ }{.}   
\LinesNumbered           
\caption{Visual Token Importance Scores}
\label{alg:vis_token_importance}
\KwIn{%
    Complete collection of multi-head attention maps from all vision encoder layers $\mathbf{A}$;
    Indices of all visual encoder layers $\mathcal{I}(L)$;
    Middle layers range parameters $m, q$; 
  
}

\KwOut{
Visual token importance scores $\mathbf{S} \in \mathbb{R}^{N \times 1}$ 
}
\SetAlgoLined

Index of selected middle layers:

\quad$L_s = \{l_m, \dots, l_{m+q}\}\subseteq \mathcal{I}(L)$;

\For{each $l \in L_s$}{
    Extract attention maps from $\mathbf{A}$: $\mathbf{A}^{(l,H)} \in \mathbb{R}^{H \times(N+1) \times (N+1)}$ \;
}
Compute importance scores:

\quad$\bar{\mathbf{A}}^{(H)} = \frac{1}{|L_s|}\sum_{l \in L_s} \mathbf{A}^{(l,H)}$\;
\quad$\bar{\mathbf{A}} = \frac{1}{H}\sum_{h=1}^{H} \bar{\mathbf{A}}^{(H)}$ \;
\quad$\mathbf{S} = \frac{1}{N+1}\sum_{j=0}^{N} \bar{\mathbf{A}}_{j,1:} $ \;
\Return{$\mathbf{S} \in \mathbb{R}^{N \times 1}$}\\
\end{algorithm}

\begin{algorithm}[t]
\footnotesize
\SetNlSty{textbf}{\ }{.}   
\LinesNumbered           
\caption{Hierarchical Visual Token Pruning}
\label{alg:hierarchical}
\KwIn{%
    Visual token importance scores $\mathbf{S} \in \mathbb{R}^{N \times 1}$;  
    Visual tokens output by penultimate vison encoder layer $\mathbf{Z}_v \in \mathbb{R}^{N \times d_v}$;
  
}

\KwOut{
Retained visual tokens $\mathbf{Z}_v^{\text{retained}} \in \mathbb{R}^{p \times d_v}$
}
\SetAlgoLined
\textit{/* Stage1 Global Retaining */} \\
Reshape :$\mathbf{Z}'_v \in \mathbb{R}^{n \times n \times d_v} \xleftarrow{} \mathbf{Z}_v \in \mathbb{R}^{N \times d_v}$\;
Region Division : $\{\mathcal{R}_i\}_{i=1}^{r^2} \xleftarrow{} \mathbf{Z}'_v$\;
\For{each region $\mathcal{R}_i$}{
  Obtain positional indices: $\mathcal{I}(\mathcal{R}_i)$ \;
  Obtain importance scores: $\mathbf{S}_i \gets \mathbf{S}[\mathcal{I}(\mathcal{R}_i)]$\;
  $\mathcal{I}^{\text{global}}_i \gets \texttt{TopK}(\mathcal{I}(\mathcal{R}_i), \mathbf{S}_i, k\%)$\;
}
$\mathcal{I}^{\text{global}} \gets \bigcup_{i=1}^{r^2} \mathcal{I}^{\text{global}}_i$\;

\vspace{0.5em}
\textit{/* Stage2 Local Retaining */} \\
$c \times c$ windowing :$\{\mathcal{W}_j\}_{j=1}^{N_w}$ $  \xleftarrow{} \mathbf{Z}'_v$ \;
\For{each window $\mathcal{W}_j$}{
  Obtain positional indices : $\mathcal{I}(\mathcal{W}_j) $\;
  Obtain candidates: $\mathcal{U}_j \gets \mathcal{I}(\mathcal{W}_j) \setminus \mathcal{I}^{\text{global}}$\;
  \If{$\mathcal{U}_j \neq \emptyset$}{
    $\mathcal{I}^{\text{local}}_j \gets \arg\max\limits_{t \in \mathcal{U}_j} \mathbf{S}[t]$\;
  }
}
$\mathcal{I}^{\text{local}} \gets \bigcup_{j=1}^{N_w} \mathcal{I}^{\text{local}}_j$\;

Final index set : $\mathcal{I}^{\text{final}} \gets \mathcal{I}^{\text{global}} \cup \mathcal{I}^{\text{local}}$ \;
Sorted final index list : $\mathcal{I}^{\text{final}}_{\text{sorted}} \gets \texttt{sort}(\mathcal{I}^{\text{final}})$ \;
Retained visual tokens : $\mathbf{Z}_v^{\text{retained}} \gets \mathbf{Z}_v[\mathcal{I}^{\text{final}}_{\text{sorted}}]$ \;

\Return{$\mathbf{Z}_v^{\text{retained}} \in \mathbb{R}^{p \times d_v}$}
\end{algorithm}

\end{document}